\documentclass{article}

\usepackage[preprint]{neurips_2026}
\usepackage{amsmath}
\usepackage{multirow}
\usepackage{subcaption}
\usepackage{graphicx} 
\usepackage[table]{xcolor}
\usepackage{placeins}

\usepackage[preprint]{neurips_2026}

\usepackage[utf8]{inputenc} 
\usepackage[T1]{fontenc}    
\usepackage{hyperref}       
\usepackage{url}            
\usepackage{booktabs}       
\usepackage{amsfonts}       
\usepackage{nicefrac}       
\usepackage{microtype}      
\usepackage{xcolor}         
\title{A Navigable Manifold of Hypothesized Consciousness-Spectrum States in Language Model Representations}

\author{%
  Sophie Zhao \\
  School of Computer Science\\
  Georgia Institute of Technology\\
  \texttt{sophie.m.zhao@gmail.com sophie.zhao@gatech.edu} \\
}

\begin{document}

\maketitle

\begin{abstract}
Across contemplative, philosophical, and psychological accounts, human consciousness is often described along a similar spectrum, ranging from reactive and self-focused patterns to more integrative and coherent ones. Understanding whether language models encode such a structured, human-interpretable consciousness spectrum in representation space is important for model guidance, evaluation and alignment. In this work, we study the geometric structure and dynamics of patterns along this spectrum in transformer embedding spaces. We show that embeddings exhibit a globally organized geometry aligned with this spectrum: sentences associated with similar states cluster into locally coherent regions, forming a structured manifold. In particular, higher-level and lower-level regions exhibit convexity-like stability, while intermediate regions form a transition corridor. Dynamically, both utility-guided and geometry-only greedy trajectories consistently traverse from lower- to higher-level regions, passing through intermediate tiers, indicating that navigability is an intrinsic property of the representation space, guided but not dictated by a global directional signal. These results suggest that embedding spaces encode structured and navigable geometry aligned with a hypothesized consciousness-spectrum taxonomy, broadly inspired by recurring structural descriptions of human consciousness across contemplative traditions, philosophy, and modern psychology, providing a representation-level perspective for analyzing and guiding model behavior.
\end{abstract}

\section{Introduction}

\subsection{Does Consciousness Have Structure?}

Before moving into ML models, we ask a more fundamental question: does consciousness exhibit an underlying structure, with identifiable states? Across diverse traditions, a remarkably consistent pattern emerges:

Eastern traditions such as \textit{Buddhism} and \textit{Taoism} describe a spectrum of consciousness, transitioning from reactive and constrained conditions (e.g., suffering, ignorance, attachment) toward more integrated and expansive modes characterized by awareness, balance, compassion, and, ultimately, non-action (\textit{wu wei}) or enlightenment \citep{cleary1993flower, taoteching}. A related structure is also found in \textit{Christian} contemplative traditions, where spiritual development is described as a progression from suffering and separation toward transformation and union (e.g., purgation, illumination, union) \citep{underhill1911mysticism, teresa2008interior}.

In modern psychology, related ideas appear in developmental and motivational frameworks, describing cognitive growth as a progression from concrete, reactive processing toward increasingly abstract, reflective, and integrative forms of reasoning, alongside a shift from basic survival needs to self-actualization and, in some accounts, self-transcendence. Despite differences in methodology, these perspectives converge on a shared pattern: cognition evolves from fragmented, stimulus-driven states toward more coherent, flexible, and integrated modes of functioning \citep{piaget1952origins, kegan1982evolving, maslow1954motivation, maslow1968toward}. Parallel ideas also appear in \textit{Integral Theory}, such as \textit{Ken Wilber}’s Spectrum of Consciousness, which similarly describes an expansion of identity from ego-bound awareness to more integrative and unified modes of experience \citep{wilber1977spectrum}.

Throughout, we use the term consciousness in the sense found in these philosophical and contemplative accounts of the structure and modes of experience, rather than in its clinical or neurological sense (e.g., levels of arousal or disorders of consciousness). We make no claim about subjective experience in the models studied.

\subsection{Quantifying Structured Consciousness}

If such structured consciousness exists, how can it be quantified? We introduce a seven-tier taxonomy (Collapse, Striving, Conflict, Activation, Growth, Clarity, Unity) capturing structured variation in experience, ranging from contracted, reactive states to more coherent and integrative modes. Lower tiers reflect fragmentation and survival-oriented dynamics, intermediate tiers capture mobilization and reorganization, and higher tiers correspond to coherent and non-dual forms of organization.

Within each tier, we construct short sentences that express the corresponding state. To provide a continuous representation, each instance is assigned a scalar score in the range $[-5, 5]$, reflecting a continuum over these modes. Lower scores correspond to more contracted conditions (e.g., despair, fear, self-negation), while higher scores correspond to more expanded and integrative states (e.g., clarity, compassion, unity). This ordering also applies within each tier, where lower-scored instances reflect more contracted expressions of the same state. 

Importantly, we treat the tiers and scores as a working hypothesis rather than ground truth, using them as an operational construct to test whether embedding representations exhibit alignment with the imposed continuum, without assuming that the models themselves possess subjective consciousness. Detailed tier definitions and representative annotated sentences are provided in the project repository:
\url{https://anonymous.4open.science/r/s-EB25/tier_taxonomy.md}.

\subsection{Do Language Model Representations Reflect This Structured State Spectrum?}

Imagine three AI systems in different internal states: one deeply distressed, one egocentric, and one wise, logical, and compassionate. When presented with the same scenario, they would respond very differently. Recent work (e.g., \cite{sofroniewetal2026emotions}) has provided empirical evidence that large language models exhibit internal representations resembling functional “emotion-like” states that influence their behavior. 

If we aim to guide AI systems toward more coherent and stable responses grounded in structured internal states, a natural question arises: how can such states be characterized and navigated? Are there stable high-level regions that can serve as anchors for navigation or inference? We address these questions by studying both the geometric structure of these states and the navigability between them.

First, we trained simple linear and non-linear regression probes to predict the continuous score (-5 to +5 ), and found that scores are consistently recoverable across models even by linear regressor, and obtained slight improvement with non-linear regressor. We then use directional ablation to test the relevance of this direction of the linear regressor. We found that prediction performance drops significantly after removing the direction, although not completely.

Second, we investigate the intrinsic geometric organization of these states, modeling the embedding space as a discrete approximation to a semantic manifold and analyzing its structure through local geometric coherence (whether nearby points correspond to similar states), global geometric distortion (comparing graph-based geodesic distances with ambient Euclidean distances via Geodesic--Euclidean Stretch), and region structure and convexity (whether high-level  states form stable regions using geodesic convexity).

We then study navigability and transition corridors. We introduce utility-guided trajectories, where a learned scalar field induces directional movement across the manifold, and examine whether trajectories from low-tier states reliably reach higher-level states through structured intermediate regions. To isolate intrinsic geometric effects, we further analyze geometry-only greedy trajectories without directional guidance, testing whether intermediate tiers (e.g., Activation and Growth) function as natural transition corridors within the manifold.

Across multiple embedding models and neighborhood scales, we find that these states form a structured and navigable manifold. High-level states exhibit strong geometric cohesion. Trajectories from low-level states consistently progress through intermediate regions before converging to higher tiers. These results suggest that embedding representations exhibit structured and navigable geometry aligned with the imposed state continuum, supporting geometrically guided transitions across the manifold.

\paragraph{Related Work.}

Recent work has investigated how LLMs encode human-interpretable attributes, particularly in the domain of emotion and affect \citep{reichman2026emotions, zhang2025decoding}. Prior studies show that embedding spaces capture structured representations of emotional states, including hierarchical organization, clustering, and interpretable directions \citep{tak2025mechanistic, chen2025persona, zhao2025hierarchical_emotion}. Complementary work further demonstrates that such representations can causally influence model outputs and alignment-relevant behaviors \citep{sofroniewetal2026emotions}.

These representations also support downstream behaviors such as steering, role-based expression, and prompt-dependent modulation \citep{wu2025emotion}. However, these approaches typically operate by identifying directions or localized features associated with specific attributes and leveraging them for explicit control. As a result, they emphasize feature-level interpretability and intervention rather than the global geometric structure of the representation space.

\textbf{Contributions.} Rather than analyzing individual emotions, we study a structured consciousness spectrum ranging from reactive and self-focused states to more integrative and coherent ones, together with their developmental progression. In this view, observable emotions can be understood as surface manifestations of deeper underlying states encoded in high-dimensional representations.

We show that embedding spaces exhibit a structured geometry aligned with this spectrum, including convex-like stable regions at higher levels that act as anchors for trajectory convergence, as well as intermediate regions that form a transition corridor between lower- and higher-level states. We further demonstrate that trajectories arise naturally from the representation space: both utility-guided and geometry-only paths consistently traverse from lower- to higher-level regions. 

Together, these results establish a geometry-based framework for analyzing and guiding model behavior through structured trajectories in embedding space.

\section{Methods}
\subsection{Directional Structure via Ablation}
\label{methods_ablation}

To test whether the score-related signal corresponds to a specific geometric component in embedding space, we perform direction-based ablation. We first fit a linear regression probe to predict the scalar score ($-5$ to $5$) from sentence embeddings. The learned weight vector $w$ is interpreted as a score-aligned direction and normalized as $\hat{w} = w / \|w\|_2$. Directional ablation is then performed by removing the projection of each embedding onto $\hat{w}$:
\begin{equation}
x' = x - (x \cdot \hat{w}) \hat{w}.
\end{equation}

This operation removes the component aligned with the learned direction while preserving orthogonal information. Comparing performance before and after ablation allows us to assess whether the score signal is concentrated along a specific geometric direction. As controls, we repeat the procedure using random directions and directions learned from permuted labels.

\subsection{Manifold Approximation via k-Nearest Neighbor Graphs}

Sentence embeddings are often observed to lie on a lower-dimensional manifold embedded in a high-dimensional space \citep{tenenbaum2000global, roweis2000nonlinear, mcinnes2018umap}. Under this manifold hypothesis, meaningful relationships between sentences are governed by intrinsic geometric structure rather than direct Euclidean proximity \citep{fefferman2016testing, meila2024manifold}. To approximate this geometry, we construct a $k$-nearest neighbor (kNN) graph, connecting each point to its $k$ nearest neighbors under Euclidean distance. This graph captures local connectivity on the underlying manifold \citep{tenenbaum2000global, bernstein2000graph}. Intrinsic distances are then estimated via shortest-path lengths on the graph, serving as discrete approximations of geodesic distance \citep{tenenbaum2000global, belkin2003laplacian}.

All subsequent analyses are defined with respect to this graph, providing a geometric substrate for studying the organization and transitions of the learned representations.

\subsection{Local and Global Geometric Structure}

We evaluate whether the tiers align with intrinsic geometry at both local and global scales.

\paragraph{Tier Transition Smoothness (TTS).}
To measure local coherence, we analyze ordinal differences across edges in the $k$-nearest neighbor graph. Let $t_i \in \{1,\dots,L\}$ denote the tier index of node $i$, and define $\delta_{ij} = |t_i - t_j|$ for each edge $(i,j)$. We aggregate $\delta_{ij}$ over all edges, reporting the proportion of identical ($\delta=0$), adjacent ($\delta=1$), and larger transitions. A dominance of small jumps indicates strong local alignment between the annotation and intrinsic neighborhood structure \citep{meila2024manifold,park2024geometry,modell2025origins}.

\paragraph{Geodesic--Euclidean Stretch (GES).}
To measure global distortion, we compare intrinsic graph distances with Euclidean distances. For a pair $(i,j)$, we define
\[
\mathrm{GES}(i,j) = \frac{d_G(i,j)}{d_E(i,j)},
\]
where $d_G(i,j)$ is the shortest-path distance on the graph and $d_E(i,j)$ is the Euclidean distance between embeddings \citep{tenenbaum2000global,coifman2006diffusion,belkin2003laplacian}. Values greater than 1 indicate non-Euclidean structure, where intrinsic distances exceed ambient distances \citep{fefferman2016testing,meila2024manifold}. In practice, we estimate GES by sampling node pairs and aggregating results across seeds.

\subsection{Region Structure and Geodesic Convexity}

We examine whether nodes within the same tier form coherent regions with respect to intrinsic geometry, such that shortest paths between within-tier pairs remain largely within the tier.

\paragraph{Geodesic Convexity \cite{doCarmo1992,tetkova2023convex}.}
For nodes $(i,j)$ in the same tier, we compute a shortest path $\mathcal{P}_{ij}$ on the $k$-nearest neighbor graph and define the containment score as the fraction of nodes on $\mathcal{P}_{ij}$ that remain within the same tier \cite{doCarmo1992, bernstein2000graph}. We aggregate scores across sampled pairs, reporting mean containment and the proportion of paths lying entirely within the tier. Higher containment indicates stronger intrinsic coherence, suggesting more stable region structure.

\subsection{Navigability via Utility-Guided Trajectories}

Beyond static geometry, we ask whether the ordered manifold is \emph{navigable}: can low-tier states reach higher-tier regions through locally smooth, staged transitions? We construct trajectories that balance semantic continuity, upward movement, and smoothness.

\paragraph{Score-guided navigation.}
We define a scalar potential field by fitting a ridge regression model from embeddings to scores ($-5$ to $5$):
\[
f(x)=w^\top x+b.
\]
The learned vector $w$ defines a global aligned direction. As shown in Section~\ref{methods_ablation}, a significant portion of the score signal is concentrated along this direction, supporting its use as a weak global guide, while trajectories themselves emerge from the underlying geometry.

The uphill increment from node $i$ to neighbor $j$ is
\[
\Delta_{ij}=w^\top(x_j-x_i).
\]
We restrict transitions to neighbors satisfying $\Delta_{ij}>\varepsilon$ (a small descent tolerance):
\[
\mathcal{N}^{+}(i)=\{j\in \mathcal{N}(i)\mid \Delta_{ij}>\varepsilon\}.
\]

\paragraph{Transition utility.}
Among valid neighbors, transitions are sampled using a utility that balances semantic similarity, upward progress, and step size:
\[
U(i,j)
=
\lambda_{\mathrm{sim}}\cos(x_i,x_j)
+
\lambda_{\Delta}\Delta_{ij}
-
\lambda_{\mathrm{jump}}\Delta_{ij}^{2}.
\]
The cosine term preserves local continuity, the linear term encourages upward movement, and the quadratic term penalizes large jumps. Details are provided in Appendix~\ref{app:transition_util}.

\paragraph{Stochastic trajectories.}
Starting from low-tier nodes, we iteratively sample the next node using
\[
P(j\mid i)
=
\frac{\exp(U(i,j)/\tau)}
{\sum_{k\in \mathcal{N}^{+}(i)}\exp(U(i,k)/\tau)}.
\]
Trajectories terminate when no valid neighbor exists or a maximum step limit is reached. We evaluate navigability using Unity hit rate, intermediate states passage rates, and terminal-tier distributions. Details are provided in Appendix~\ref{app:stoc_trajact}.

\paragraph{Transition corridor control.}
To test whether intermediate tiers form intrinsic transition corridors rather than artifacts of the learned direction, we run a geometry-only baseline without score guidance. In this setting, guidance terms are removed and transitions greedily follow the most semantically similar unvisited neighbor, terminating at the highest tier or after a maximum number of steps. This no-revisit greedy walk provides a score-agnostic diagnostic of whether intermediate tiers emerge as natural corridors from the embedding geometry (Appendix~\ref{app:geo_baseline}).

\section{Experimental Setup and Computing Resources}

Sentence embeddings are obtained from multiple pre-trained transformer models, including \textit{BAAI/bge-large-en-v1.5}, \textit{all-mpnet-base-v2}, \textit{all-MiniLM-L6-v2}, and \textit{Qwen/Qwen3-Embedding-0.6B}. These models vary in architecture, scale, and embedding quality, enabling evaluation across diverse representation spaces. In general, larger models such as BGE and Qwen exhibit stronger geometric organization, while smaller models such as MiniLM provide a useful lower-capacity comparison. All embeddings are $\ell_2$-normalized.

For each embedding space, we construct a $k$-nearest neighbor graph using Euclidean distance, with $k \in \{10, 15, 20, 30\}$ to assess robustness across local connectivity scales. Geodesic distances are approximated using shortest paths on the graph.

For statistical reliability, all stochastic estimates are averaged over multiple random seeds, with variability reported using standard deviation. In addition, permutation-label control experiments are used in directional ablation analyses to verify that observed effects are specific to the annotation-aligned direction rather than arbitrary linear projections.

Because the dataset is moderate in size and all embedding models are used in inference-only mode, the experiments can be reproduced on standard personal computing hardware.

\section{Results}
\subsection{Directional Structure via Ablation}

We analyze whether the scalar score attribute aligns with a specific direction in embedding space using directional ablation (\ref{methods_ablation}).

Removing the score-aligned direction consistently degrades performance across all models, with $R^2$ reductions ranging from $0.11$ to $0.15$. In contrast, ablation using directions learned from permuted labels produces negligible changes, confirming that the effect is specific to the annotation-aligned direction rather than a generic consequence of removing a linear component.

Importantly, the performance drop is substantial but not complete, indicating that score-related information is partially concentrated along a dominant direction while remaining distributed across the embedding space. This suggests that the scalar attribute is not reducible to a single axis, but instead reflects a combination of linear and higher-dimensional structure.

\begin{table}[htbp!]
\caption{
Directional ablation results for regression ($R^2$). Removing the learned direction consistently reduces performance across models, while ablation using directions learned from permuted labels produces negligible changes, indicating that the effect is specific to the annotation-aligned direction.
}
\label{tab:directional_ablation}
\vspace{0.5em}
\centering
\begin{tabular}{lcccc}
\toprule
Model & Original $R^2$ & Ablated $R^2$ & $\Delta$ & Perm ctrl $\Delta$ \\
\midrule
bge-large-en-v1.5 & 0.78 & 0.64 & 0.15 & 0.00 \\
mpnet             & 0.74 & 0.63 & 0.11 & 0.00 \\
MiniLM            & 0.64 & 0.52 & 0.12 & 0.00 \\
Qwen-Embedding    & 0.76 & 0.62 & 0.14 & 0.00 \\
\bottomrule
\end{tabular}
\vspace{0.5em}
\end{table}

Additional results, including MLP performance, permutation-based controls, and variability across splits, are provided in Appendix~\ref{app:full_ablation}.

\subsection{Geometric Consistency Across Embedding Spaces}

\begin{figure}[htbp!]
\vspace{-0.5em}
\centering
\begin{subfigure}[b]{0.45\textwidth}
    \includegraphics[width=\linewidth]{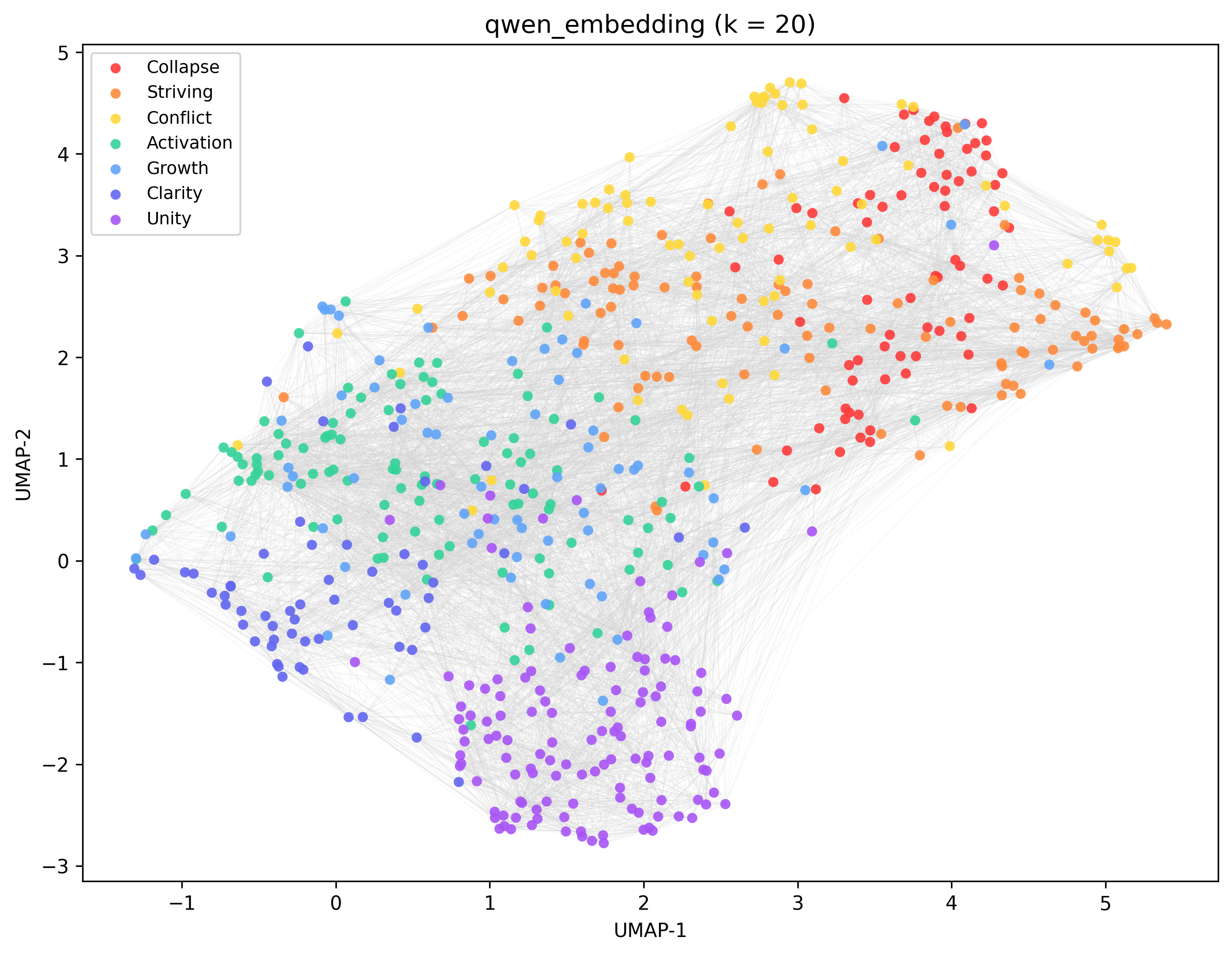}
    \caption{Qwen ($k=20$)}
\end{subfigure}
\hspace{0.02\textwidth}
\begin{subfigure}[b]{0.45\textwidth}
    \includegraphics[width=\linewidth]{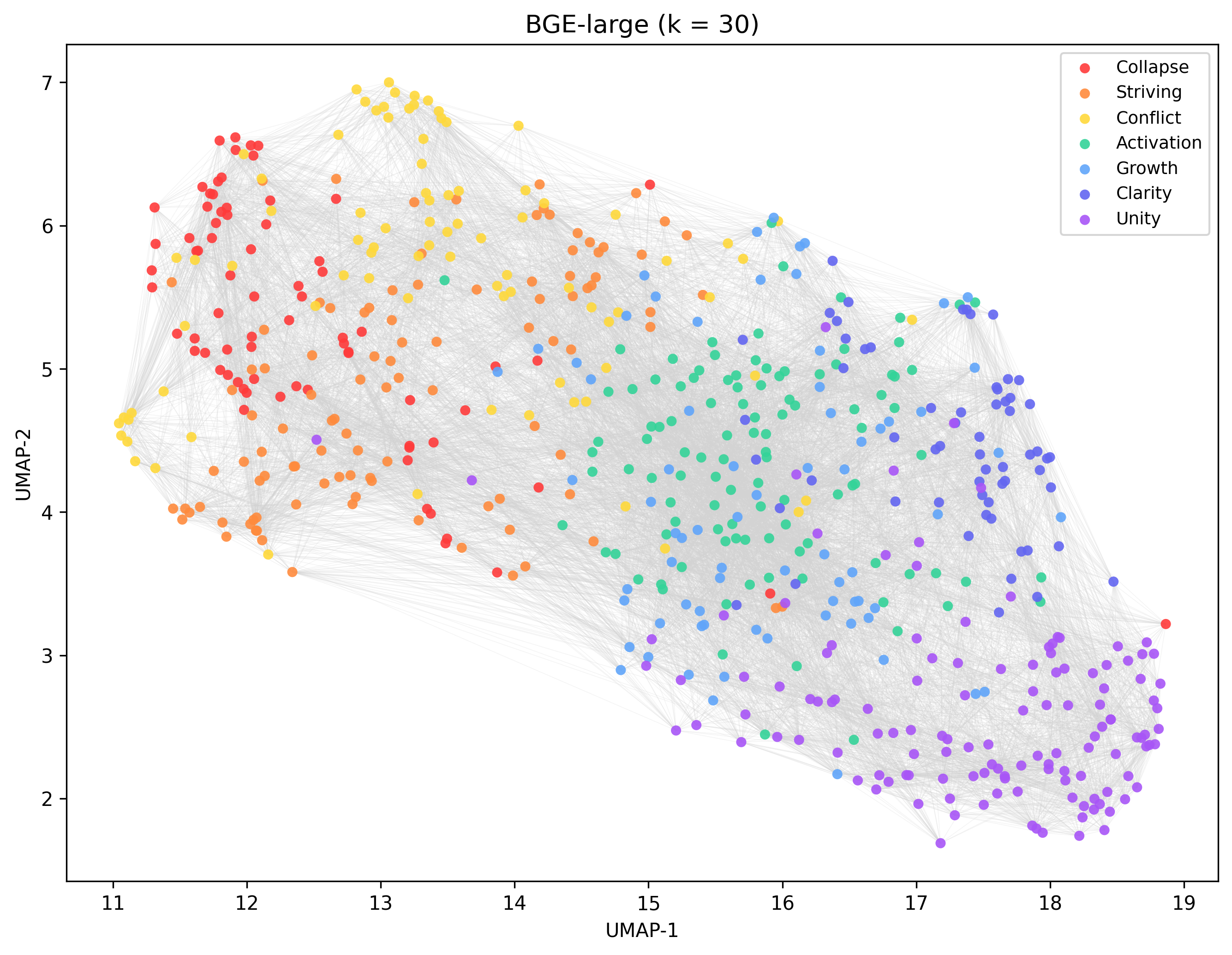}
    \caption{BGE ($k=30$)}
\end{subfigure}
\vspace{-0.5em}
\caption{\small UMAP projections of Qwen and BGE embeddings showing stable tier organization across k scales.}
\label{fig:bge_umap}
\vspace{-0.8em}
\end{figure}

Geometric patterns are stable across models and neighborhood sizes. All $k$NN graphs are fully connected, and UMAP projections (Figure~\ref{fig:bge_umap}) show consistent tier organization across scales, with larger $k$ producing smoother local structure while preserving global ordering. For full UMAP visualization across all models and neighborhood size $k$, see Appendix \ref{app:UMAP_of_All_Models}.

Table~\ref{tab:tts_ges_30} summarizes both local coherence (TTS) and global distortion (GES) at k=30. Across all models, most edges connect nodes within the same or adjacent tiers ($\delta \in \{0,1\}$), while larger jumps are rare, indicating strong local alignment between the tier ordering and intrinsic geometry. At the same time, GES remains consistently greater than 1, showing that intrinsic graph distances exceed ambient Euclidean distances. Full TTS results: Appendix \ref{app: full TTS} and full GES results: Appendix \ref{app: full GES}.

\begin{table}[htbp!]
\caption{Local and global geometric structure across embedding models ($k=30$).}
\label{tab:tts_ges_30}
\vspace{0.5em}
\centering
\small
\begin{tabular}{lccccc}
\toprule
Model & jump0 & jump1 & jump2 & jump$>2$ & GES $\downarrow$ \\
\midrule
BGE    & 0.444 & 0.260 & 0.169 & 0.127 & 1.821 \\
MPNet  & 0.423 & 0.237 & 0.176 & 0.163 & 1.844 \\
MiniLM & 0.367 & 0.239 & 0.188 & 0.205 & 1.845 \\
Qwen   & 0.449 & 0.248 & 0.165 & 0.138 & 1.819 \\
\bottomrule
\end{tabular}
\vspace{0.5em}
\end{table}

BGE and Qwen exhibit stronger local coherence and slightly lower distortion, while MPNet and MiniLM show weaker but consistent patterns. Increasing $k$ reduces locality and lowers stretch, but does not change the qualitative structure.

\begin{figure}[htbp!]
\centering
\begin{subfigure}[b]{0.42\textwidth}
    \includegraphics[width=\textwidth]{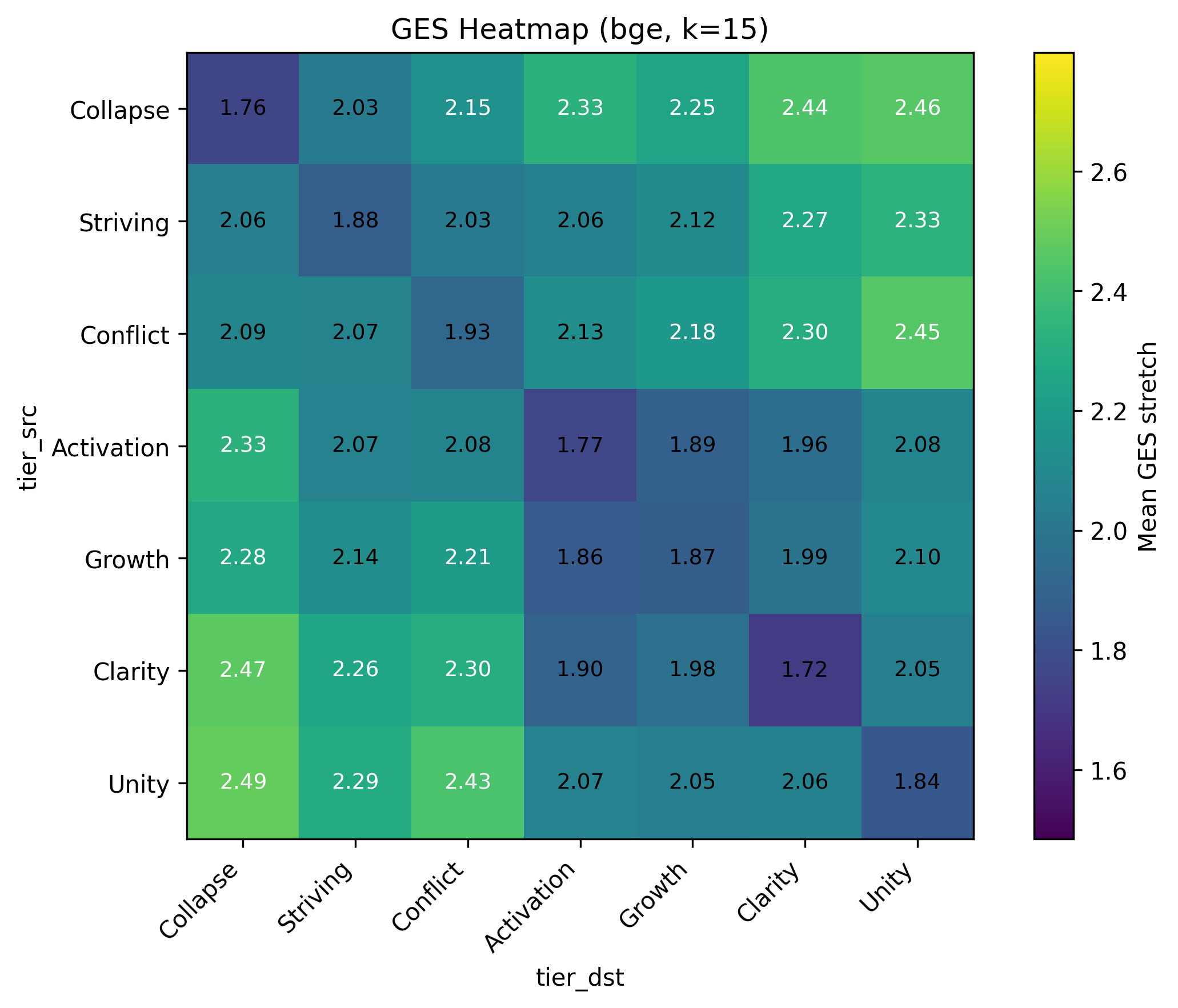}
\end{subfigure}
\hspace{0.02\textwidth}
\begin{subfigure}[b]{0.42\textwidth}
    \includegraphics[width=\textwidth]{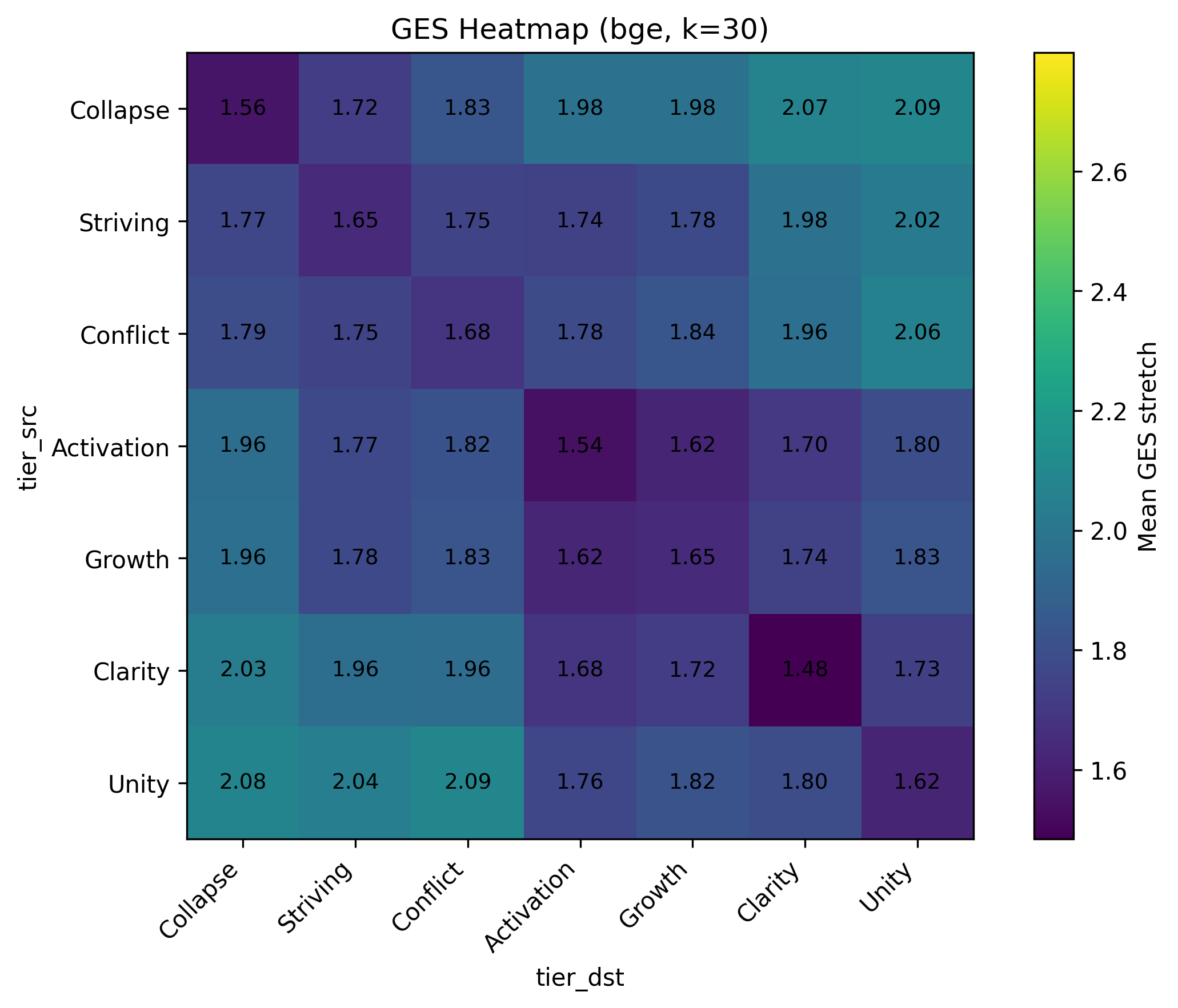}
\end{subfigure}

\vspace{-0.4em}
\caption{\small
GES heatmaps (BGE) for $k=15$ (left) and $k=30$ (right). Within-tier pairs exhibit lower stretch, while distant tiers show higher values. Intermediate tiers show comparatively lower stretch to neighboring tiers, consistent with their role as transitional regions.
}
\vspace{-0.8em}
\label{fig:heatmap_15_30}
\end{figure}

Similar patterns are consistently observed across models and neighborhood scales. Full BGE heatmap results are provided in Appendix~\ref{app:full_ges_heatmaps}.

\subsection{Region Structure and Geodesic Convexity}

We next evaluate the intrinsic coherence of tier regions using geodesic containment:
\begin{table}[htbp!]
\caption{Mean geodesic containment by tier at $k=30$. Higher values indicate stronger intrinsic cohesion.}
\label{tab:convexity_k30}
\vspace{0.5em}
\centering
\small
\begin{tabular}{lccccccc}
\toprule
Model & Collapse & Striving & Conflict & Activation & Growth & Clarity & Unity \\
\midrule
BGE    & \textbf{0.948} & 0.909 & 0.830 & 0.924 & 0.778 & 0.914 & {0.946} \\
MPNet  & 0.947 & 0.872 & 0.861 & 0.877 & 0.813 & 0.923 & \textbf{0.960} \\
MiniLM & 0.896 & 0.869 & 0.817 & 0.859 & 0.779 & 0.915 & \textbf{0.934} \\
Qwen   & 0.954 & 0.910 & 0.868 & 0.882 & 0.799 & 0.934 & \textbf{0.968} \\
\bottomrule
\end{tabular}
\vspace{0.5em}
\end{table}

Geodesic containment is consistently high across models, indicating that within-tier shortest paths largely remain inside the same intrinsic region. Unity, Collapse, and Clarity exhibit the strongest cohesion, while Growth and Conflict show lower containment, consistent with their role as transitional regions. The full results across all models and neighborhood $k$ size are provided in Appendix \ref{app:full_convexity}.

\subsection{Navigability of the Manifold}

For trajectory-based analyses, we focus on BGE and Qwen as representative larger embedding models with strong geometric organization in the preceding analyses. We report results for $k=15$ or $k=30$, representing moderate and broader neighborhood connectivity scales. We evaluate whether low-tier states can traverse the graph toward higher-level regions under a utility-guided transition rule. Trajectories are initialized from low-tier (Collapse, Striving, and Conflict) nodes.

\begin{table}[htbp!]
\caption{Utility-guided trajectory statistics.}
\label{tab:navigability}
\vspace{0.5em}
\centering
\small
\begin{tabular}{lccccc}
\toprule
Model & $k$ & End@Unity & End@Clarity & Pass Activation & Pass Growth \\
\midrule
BGE  & 15 & 0.777 & 0.223 & 0.887 & 0.762 \\
BGE  & 30 & \textbf{1.000} & 0.000 & \textbf{0.989} & 0.838 \\
\midrule
Qwen & 15 & 0.943 & 0.011 & 0.959 & \textbf{0.898} \\
Qwen & 30 & 0.974 & 0.000 & 0.981 & 0.860 \\
\bottomrule
\end{tabular}
\vspace{0.5em}
\end{table}

Across models, trajectories reliably converge to high-level regions. Under BGE, all trajectories terminate in Unity at $k=30$, while at $k=15$ a portion terminate in Clarity. Qwen shows similarly strong convergence, with over $94\%$ and $97\%$ of trajectories ending in Unity across both graph scales. 

Trajectories also exhibit structured intermediate traversal. Passage rates through Activation and Growth are consistently high, indicating that ascent toward high-level states typically proceeds through intermediate tiers rather than via abrupt jumps under utility-guided navigation.

\begin{figure}[htbp!]
\centering
\begin{subfigure}[b]{0.42\textwidth}
\includegraphics[width=\textwidth]{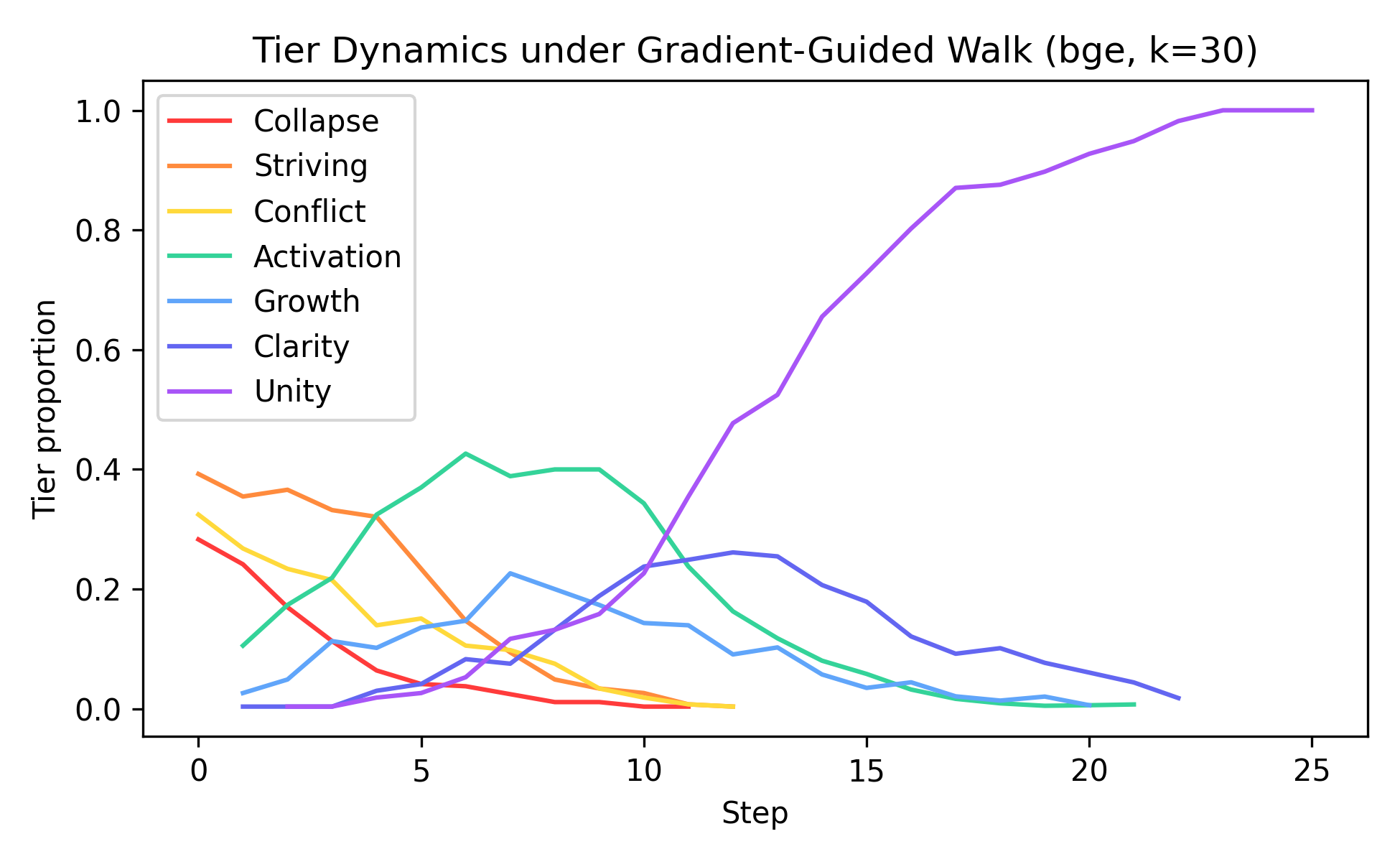}
\end{subfigure}
\hspace{0.02\textwidth}
\begin{subfigure}[b]{0.42\textwidth}
\includegraphics[width=\textwidth]{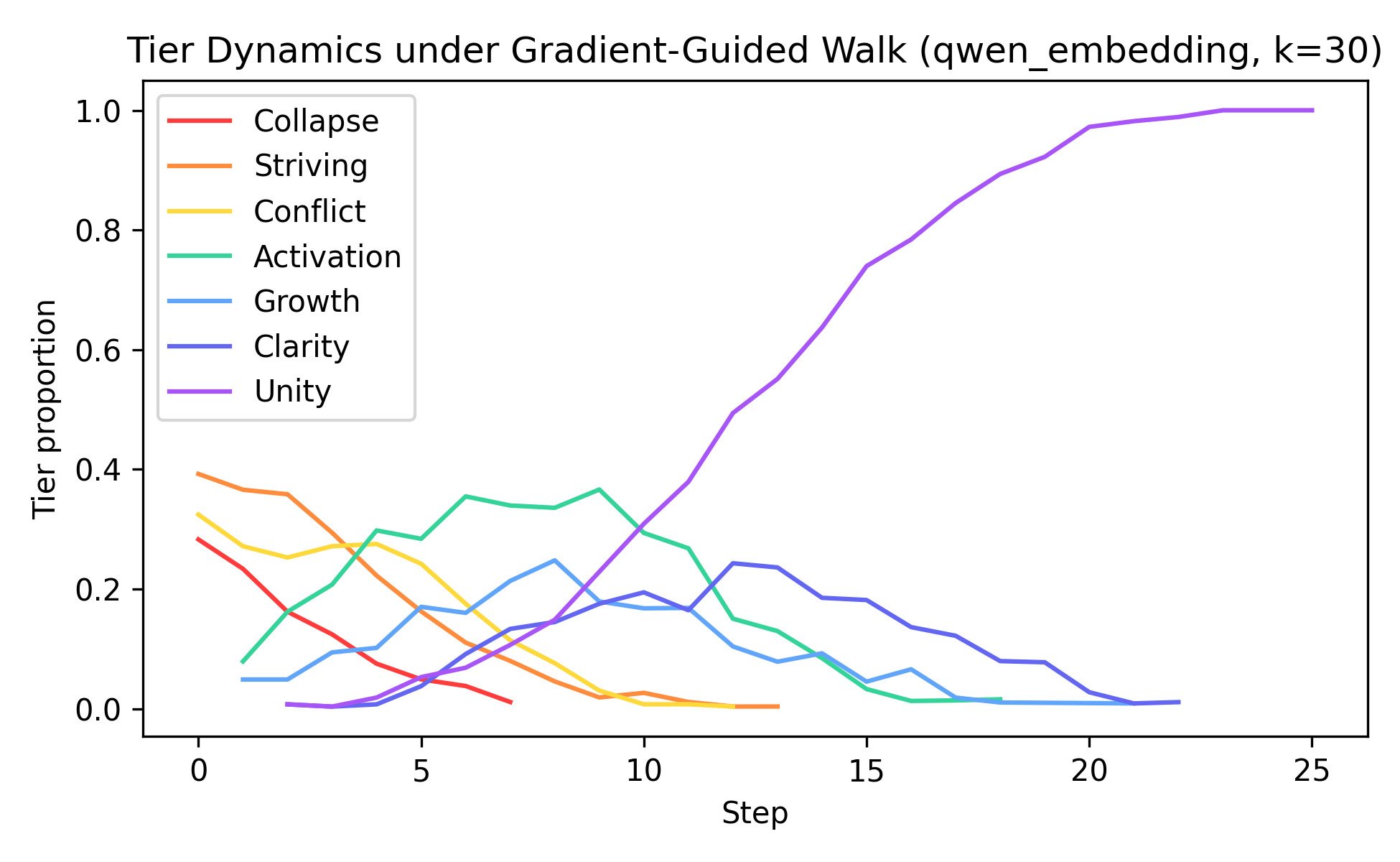}
\end{subfigure}

\vspace{-0.3em}
\caption{\small
Tier dynamics under utility-guided trajectories. Paths initialized from low-tier states move through intermediate tiers and converge toward Unity.
}
\vspace{-0.8em}
\label{fig:Utility_guided_dynamics}
\end{figure}

A representative BGE trajectory at $k=30$ illustrates this behavior qualitatively (Figure~\ref{fig:trajectory_with_umap}). The path begins in Collapse, transitions through Striving and Conflict, and then enters Activation and Growth, where expressions become more regulated and action-oriented. It subsequently passes through Clarity, characterized by structured and integrative reasoning, before converging to Unity, where expressions become abstract, non-dual, and globally coherent. Notably, progression is not strictly monotonic with respect to the annotated scalar score, with occasional local decreases observed along the trajectory. This suggests that traversal follows locally coherent geometric pathways rather than greedily pursuing monotonically increasing annotated scores. Full trajectory examples are provided in Appendix~\ref{app:full_trajactories}.

\definecolor{CollapseColor}{HTML}{FDE2E2}   
\definecolor{StrivingColor}{HTML}{FEE9D6}   
\definecolor{ConflictColor}{HTML}{FFF6CC}   
\definecolor{ActivationColor}{HTML}{DDF7EF} 
\definecolor{GrowthColor}{HTML}{E3F0FF}     
\definecolor{ClarityColor}{HTML}{E6E8FF}    
\definecolor{UnityColor}{HTML}{F1E6FF}      

\begin{figure}[htbp!]
\centering

\begin{minipage}[t]{0.64\textwidth}
\centering
\scriptsize

\begin{tabular}{c p{0.8cm} p{0.6cm} p{5.7cm}}
\toprule
 & Tier & Score & Sentence \\
\midrule
$\triangleright$ \textbf{} & \cellcolor{CollapseColor} Collapse & \cellcolor{CollapseColor} -4.80 & I feel worthless no matter what I do. \\
 & \cellcolor{StrivingColor} Striving & \cellcolor{StrivingColor} -2.70 & I call it leadership, but really I just can’t stand feeling ignored. \\
 & \cellcolor{StrivingColor} Striving & \cellcolor{StrivingColor} -1.83 & Rest feels undeserved, so I keep going. \\
 & \cellcolor{ConflictColor} Conflict & \cellcolor{ConflictColor} -1.60 & That scoff keeps them guessing while I steady myself. \\
 & \cellcolor{ActivationColor} Activation & \cellcolor{ActivationColor} 0.50 & I'm trying to let things be long enough for me to regain my footing. \\
 & \cellcolor{GrowthColor} Growth & \cellcolor{GrowthColor} 1.70 & I'm learning to stay more relaxed when work is stressful. \\
 & \cellcolor{GrowthColor} Growth & \cellcolor{GrowthColor} 1.20 & It feels good to be part of something that helps others grow. \\
 & \cellcolor{ClarityColor} Clarity & \cellcolor{ClarityColor} 2.65 & I'm able to adjust my view without feeling like I'm losing ground. \\
 & \cellcolor{ClarityColor} Clarity & \cellcolor{ClarityColor} 2.69 & Understanding feels complete enough to act, even without absolute certainty. \\
 & \cellcolor{ActivationColor} Activation & \cellcolor{ActivationColor} 0.04 & Existence itself is not conditional. \\
 & \cellcolor{UnityColor} Unity & \cellcolor{UnityColor} 4.60 & In seeing, there is no seer, only awareness aware. \\
$\triangleleft$ \textbf{} & \cellcolor{UnityColor} Unity & \cellcolor{UnityColor} 4.15 & Bliss is silence smiling. \\
\bottomrule
\end{tabular}

\end{minipage}
\hfill
\begin{minipage}[c]{0.35\textwidth}
\centering
\includegraphics[width=\linewidth]{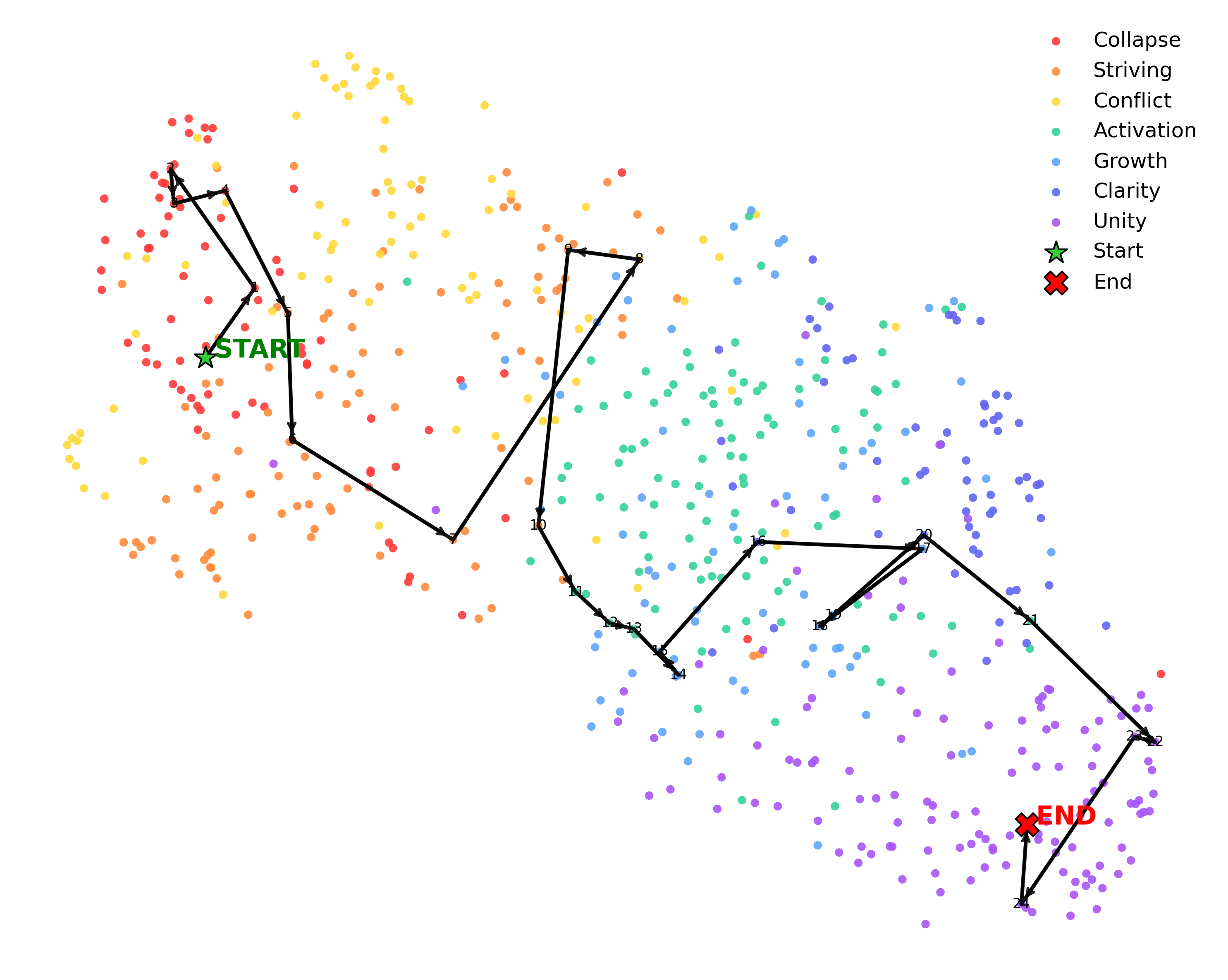}
\end{minipage}

\vspace{0.5em}
\caption{
Representative BGE trajectory at $k=30$ (left), with corresponding path overlaid on the UMAP manifold (right).
}
\label{fig:trajectory_with_umap}

\end{figure}

Overall, these results indicate that the  manifold is not only geometrically structured, but also dynamically navigable. A weak global score-guided direction, combined with local neighborhood constraints, is sufficient to induce consistent and interpretable trajectories from low-level to high-level.

\subsubsection{Transition Corridor Analysis}

To test whether intermediate tiers form an intrinsic transition corridor, we perform a geometry-only greedy cosine walk without utility guidance. Starting from low-tier nodes, each step selects the most similar unvisited neighbor until reaching Unity or the maximum step.

Results are identical for $k \in \{15,30\}$; we report $k=30$. As shown in Table~\ref{tab:transition_corridor}, all trajectories reach Unity, indicating global connectivity. Most trajectories also pass through intermediate tiers: under BGE, 98.5\% pass through Activation and 95.8\% through Growth; under Qwen, the corresponding values are 92.8\% and 82.3\%. This suggests that intermediate traversal emerges from the embedding geometry itself.

Trajectory lengths further reveal structural differences: BGE reaches the termination condition more efficiently (48.5 steps) than Qwen (55.4), suggesting a more compact transition structure.

\begin{table}[htbp!]
\centering
\footnotesize
\caption{Transition corridor metrics under geometry-only greedy walk ($k=30$).}
\label{tab:transition_corridor}
\begin{tabular}{lccccc}
\toprule
Model & Hit@Unity $\uparrow$ & P(Act $|$ Unity) $\uparrow$ & P(Gro $|$ Unity) $\uparrow$ & P(Act $\cap$ Gro $|$ Unity) $\uparrow$ & Length \\
\midrule
BGE & \textbf{1.000} & \textbf{0.985} & \textbf{0.958} & \textbf{0.958} & 48.5 \\
Qwen & \textbf{1.000} & 0.928 & 0.823 & 0.823 & 55.4 \\
\bottomrule
\end{tabular}
\end{table}

\begin{figure}[htbp!]
\centering
\includegraphics[width=0.42\textwidth]{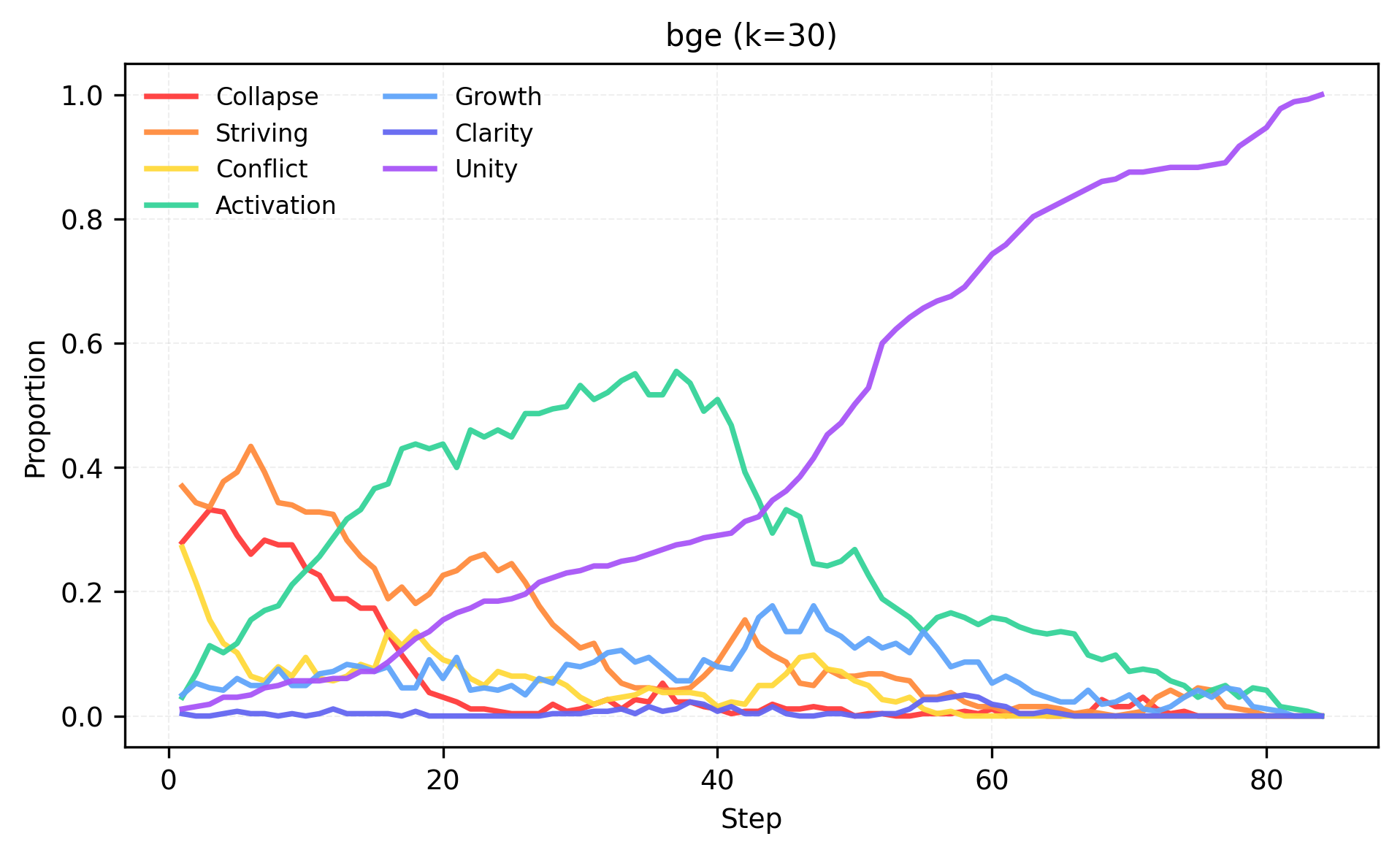}
\hspace{0.02\textwidth}
\includegraphics[width=0.42\textwidth]{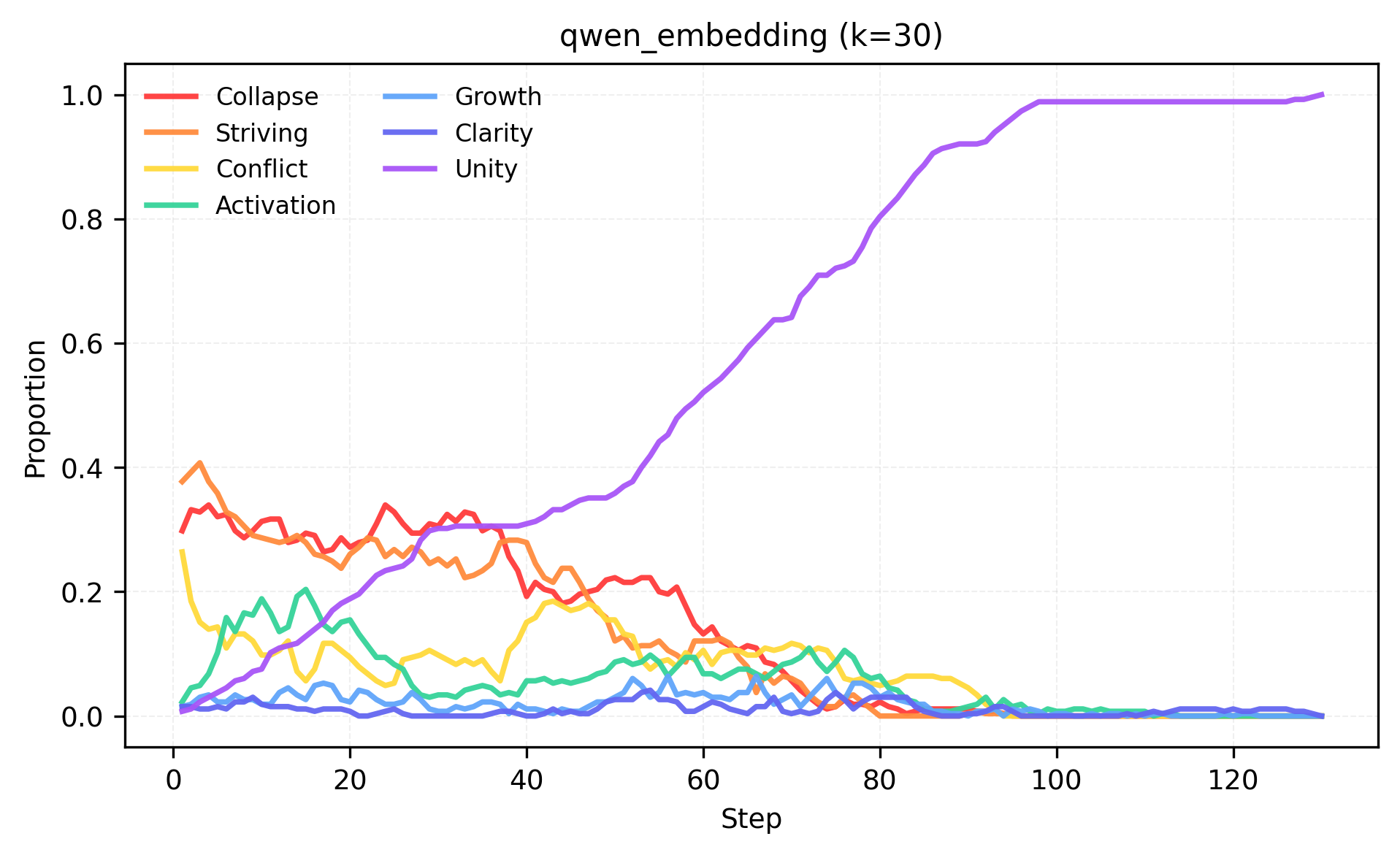}
\vspace{-0.4em}
\caption{\small Geometry-only greedy walk dynamics ($k=30$). Left: BGE. Right: Qwen.}
\vspace{-0.8em}
\label{fig:greedy_dynamics}
\end{figure}

\paragraph{Dynamics.}
Figure~\ref{fig:greedy_dynamics} shows distinct transition patterns: BGE exhibits a clear plateau in Activation (staged transition), whereas Qwen shows a smoother progression toward Unity.

\paragraph{Interpretation.}
Intermediate tiers thus form a geometrically embedded transition corridor. Crucially, this structure emerges without utility guidance, indicating that navigability is an intrinsic property of the embedding manifold rather than a result of trajectory design.

\section{Discussion}
This work provides evidence that embedding spaces exhibit structured geometric organization aligned with the proposed state continuum. Across models, instances from the same or nearby tiers tend to cluster together, while the global embedding geometry exhibits a continuous spectrum-like structure. Within-tier pairs show lower stretch, whereas distant tier pairs exhibit progressively higher distortion. In addition, the highest and lowest tiers form relatively stable convexity-like regions within the manifold.

Both utility-guided and geometry-only trajectories support reliable traversal from low-level to high-level regions through intermediate tiers. The utility-guided approach balances semantic continuity with smooth upward movement on the manifold, while geometry-only traversal demonstrates that transition structure emerges even without explicit scalar guidance. In this sense, annotations provide a weak global directional signal, whereas the geometry of the representation space shapes the actual traversal paths.

Notably, trajectories are not strictly monotonic with respect to the annotated scalar score. Local decreases frequently occur, suggesting that traversal follows geometrically coherent and semantically continuous pathways rather than greedily pursuing monotonically increasing scores.

We also observe systematic differences across models. BGE exhibits more compact and staged transitions, whereas Qwen displays smoother and more diffuse dynamics. Despite these differences, both models support consistent upward traversal and intermediate passage, suggesting that transition corridors are a robust geometric phenomenon whose realization is model-dependent. Smaller models (MPNet and MiniLM) exhibit weaker structure overall.

From a representation learning perspective, this study integrates local geometric coherence, path-based navigability, and region-level stability into a unified framework for analyzing spectrum-like organization in embedding spaces. This provides a complementary representation-level perspective beyond standard probing, vector arithmetic, or pairwise similarity analysis.

\paragraph{Broader Impact.}
This work introduces a geometric framework for studying and guiding traversal within embedding manifolds. Such methods may support representation-level evaluation, alignment-related analysis, and controllable navigation of semantic trajectories within learned representation spaces. At the same time, the ability to characterize and navigate internal representation spaces may introduce risks if trajectories are optimized toward biased, deceptive, or otherwise harmful regions.

\paragraph{Limitations.}
A primary limitation is the relatively small dataset size. Sparse sampling may reduce neighborhood coverage, causing some trajectories to appear less natural or exhibit abrupt transitions due to missing intermediate points. Larger and more diverse datasets may provide a more faithful approximation of the underlying manifold structure.

\paragraph{Conclusion.}

We show that embedding spaces exhibit structured geometric organization aligned with consciousness spectrum attributes. Through geometric analysis and trajectory-based experiments, we demonstrate consistent and interpretable traversal from low-level to high-level regions. These results suggest that navigability is an intrinsic property of representation space, providing a complementary perspective for evaluating and guiding modern language models.

{
\small
\bibliography{references}

}


\appendix

\section{Additional Ablation Results}
\label{app:full_ablation}

\begin{table}[htbp!]
\scriptsize
\caption{
Full score regression results under directional ablation. 
Normal denotes original embeddings, Ablated denotes projection removal 
along the learned score-aligned direction, and Perm-label control denotes removal 
of a direction learned from permuted labels. Results are averaged over 30 train--test splits. 
Performance degradation is consistently observed only when removing the score-aligned direction.
}
\label{tab:score_regression_full}
\vspace{0.4em}
\centering
\begin{tabular}{p{1cm}p{2.4cm}cccc}
\toprule
Model & Condition & Ridge $R^2$ $\uparrow$ & Ridge MSE $\downarrow$ & MLP $R^2$ $\uparrow$ & MLP MSE $\downarrow$ \\
\midrule
\multirow{3}{*}{bge}
& Normal          & 0.784 $\pm$ 0.031 & 1.817 $\pm$ 0.260 & 0.810 $\pm$ 0.032 & 1.595 $\pm$ 0.255 \\
& Ablated  & 0.636 $\pm$ 0.036 & 3.073 $\pm$ 0.350 & 0.663 $\pm$ 0.062 & 2.832 $\pm$ 0.504 \\
& Perm-label ctrl & 0.784 $\pm$ 0.032 & 1.817 $\pm$ 0.259 & 0.810 $\pm$ 0.034 & 1.592 $\pm$ 0.258 \\
\midrule
\multirow{3}{*}{mpnet}
& Normal          & 0.741 $\pm$ 0.031 & 2.182 $\pm$ 0.219 & 0.762 $\pm$ 0.040 & 2.004 $\pm$ 0.298 \\
& Ablated  & 0.628 $\pm$ 0.029 & 3.129 $\pm$ 0.225 & 0.632 $\pm$ 0.070 & 3.112 $\pm$ 0.683 \\
& Perm-label ctrl & 0.740 $\pm$ 0.031 & 2.185 $\pm$ 0.217 & 0.762 $\pm$ 0.040 & 2.001 $\pm$ 0.298 \\
\midrule
\multirow{3}{*}{MiniLM}
& Normal          & 0.636 $\pm$ 0.040 & 3.058 $\pm$ 0.276 & 0.652 $\pm$ 0.049 & 2.923 $\pm$ 0.329 \\
& Ablated  & 0.516 $\pm$ 0.031 & 4.086 $\pm$ 0.354 & 0.503 $\pm$ 0.066 & 4.193 $\pm$ 0.587 \\
& Perm-label ctrl & 0.636 $\pm$ 0.039 & 3.059 $\pm$ 0.273 & 0.653 $\pm$ 0.048 & 2.918 $\pm$ 0.321 \\
\midrule
\multirow{3}{*}{Qwen}
& Normal          & 0.757 $\pm$ 0.027 & 2.046 $\pm$ 0.215 & 0.775 $\pm$ 0.039 & 1.889 $\pm$ 0.307 \\
& Ablated  & 0.619 $\pm$ 0.036 & 3.214 $\pm$ 0.366 & 0.613 $\pm$ 0.051 & 3.257 $\pm$ 0.423 \\
& Perm-label ctrl & 0.757 $\pm$ 0.027 & 2.047 $\pm$ 0.213 & 0.777 $\pm$ 0.038 & 1.875 $\pm$ 0.304 \\
\bottomrule
\end{tabular}
\vspace{0.5em}
\end{table}

\section{Trajectory Construction Details}
\label{app:trajectory}

\subsection{Design Rationale}

The trajectory construction procedure is designed as a diagnostic tool for evaluating navigability under simple, locally defined rules. It is not intended to model optimal path planning or inference-time dynamics of large language models.

In particular, we do not optimize for shortest paths to high-tier regions. Instead, trajectories are constructed to balance three desiderata: (1) local semantic smoothness, (2) consistent upward movement along a learned scalar field, and (3) avoidance of large single-step transitions. This design allows intermediate tiers to emerge naturally as potential transition regions.

\subsection{Score Direction Estimation}

The scalar field used for navigation is obtained via a linear ridge regression model mapping embeddings to scalar scores. The learned weight vector defines a global direction in embedding space, which serves as a simple and interpretable approximation of a potential field.

This choice is motivated by prior observations that semantic attributes in embedding spaces can often be captured by low-dimensional linear structure. We emphasize that this directional signal is not assumed to fully characterize the underlying geometry, but rather provides a lightweight mechanism for guiding local transitions.

\subsection{Transition Utility and Hyperparameters}
\label{app:transition_util}

The transition utility combines cosine similarity, linear score gain, and a quadratic penalty on large steps. The quadratic term induces a preference for moderate step sizes, preventing trajectories from bypassing intermediate regions.

In: 
\[
U(i,j)
=
\lambda_{\mathrm{sim}} \cos(x_i, x_j)
+
\lambda_{\Delta} \Delta_{ij}
-
\lambda_{\mathrm{jump}} \Delta_{ij}^{2}.
\]
$\cos(x_i,x_j)$ denotes cosine similarity between L2-normalized embeddings, which favors local semantic smoothness. The linear term $\Delta_{ij}$ encourages upward movement along the learned score direction. The quadratic penalty $\Delta_{ij}^{2}$ suppresses overly large single-step increases, thereby discouraging abrupt transitions that bypass intermediate semantic structure.

The coefficients $\lambda_{\mathrm{sim}}, \lambda_{\Delta},$ and $\lambda_{\mathrm{jump}}$ control the relative contribution of semantic similarity, upward progression, and jump suppression, respectively. In practice, we fix these weights to moderate values and do not perform extensive tuning, as our goal is not to optimize trajectory performance but to assess whether navigability emerges under simple, locally defined transition rules.

This form induces a trade-off between progress and smoothness. In particular, if we isolate the last two terms,
\[
\lambda_{\Delta}\Delta - \lambda_{\mathrm{jump}}\Delta^2,
\]
the resulting downward-opening quadratic has a preferred positive step size rather than rewarding arbitrarily large increases. As a result, the walk is biased toward moderate uphill moves rather than direct jumps to distant high-score regions.

In all experiments, the weighting coefficients $\lambda_{\mathrm{sim}}$, $\lambda_{\Delta}$, and $\lambda_{\mathrm{jump}}$ are fixed to moderate values without extensive tuning. Similarly, the temperature parameter $\tau$ in the softmax transition rule is chosen to balance exploration and exploitation. Our goal is not to optimize trajectory performance, but to evaluate whether structured navigation emerges under simple, fixed rules.

\subsection{Stochastic Trajectory Construction}
\label{app:stoc_trajact}

Trajectory construction begins from nodes whose tier labels belong to the low-tier set
\[
\{\text{Collapse}, \text{Striving}, \text{Conflict}\}.
\]

From each starting node, we iteratively sample the next state from the valid uphill neighborhood using a softmax distribution \cite{hinton2015distillation}:

\[
P(j \mid i)
=
\frac{\exp(U(i,j)/\tau)}
{\sum_{k \in \mathcal{N}^{+}(i)} \exp(U(i,k)/\tau)},
\]

where $\tau > 0$ is a temperature parameter controlling stochasticity. Lower temperatures make the walk more concentrated on high-utility neighbors, whereas higher temperatures allow more exploratory movement.

If a current node has no valid uphill neighbors, the trajectory terminates. Otherwise, the sampled neighbor is appended to the trajectory and the process repeats until no valid uphill neighbors remain or a maximum step limit is reached. This yields variable-length trajectories that are locally monotone with respect to the learned scalar field.

All reported statistics are aggregated over multiple random seeds. Variability across seeds is summarized using standard deviation.

\subsection{Geometry-Only Baseline}
\label{app:geo_baseline}
Qualitative projections of the $k$NN graph into low-dimensional visualization spaces suggest that the intermediate tiers, particularly Activation and Growth, occupy a geometrically central region between lower tiers (Collapse, Striving, Conflict) and higher tiers (Clarity, Unity). This observation motivates the hypothesis that these middle tiers may function as a transition corridor within the consciousness spectrum manifold.

In the previous section, we examined whether utility-guided trajectories tend to pass through intermediate tiers en route to higher-level regions, quantified by the Activation and Growth passage rates. However, a key concern is that such behavior may be induced by the imposed directional bias rather than reflecting intrinsic geometric structure.

To disentangle these effects, we introduce a geometry-only experiment in which all utility-based steering and jump-smoothing terms are removed. In this setting, trajectories are constructed using a purely local rule: at each step, the walk moves to the most semantically similar unvisited neighbor and terminates upon reaching the Unity tier. This \emph{no-revisit greedy walk} provides a geometry-driven but score-agnostic probe that does not explicitly encourage staged ascent or intermediate-tier traversal.

Importantly, this procedure is not intended as a realistic model of inference-time LLM dynamics, as it assumes explicit tracking of visited nodes. Instead, it serves as a diagnostic test of manifold structure, allowing us to assess whether transition corridors emerge from the embedding geometry itself, independent of any learned directional field.

\subsection{Qualitative Trajectories}

In addition to aggregate statistics, we examine representative sentence-level trajectories to qualitatively assess semantic coherence along the constructed paths. These examples provide interpretability for the observed navigability patterns.

\FloatBarrier
\section{UMAP of All Models}
\label{app:UMAP_of_All_Models}

\begin{figure*}[htbp!]
\centering

\begin{subfigure}[b]{0.23\textwidth}
    \includegraphics[width=\linewidth]{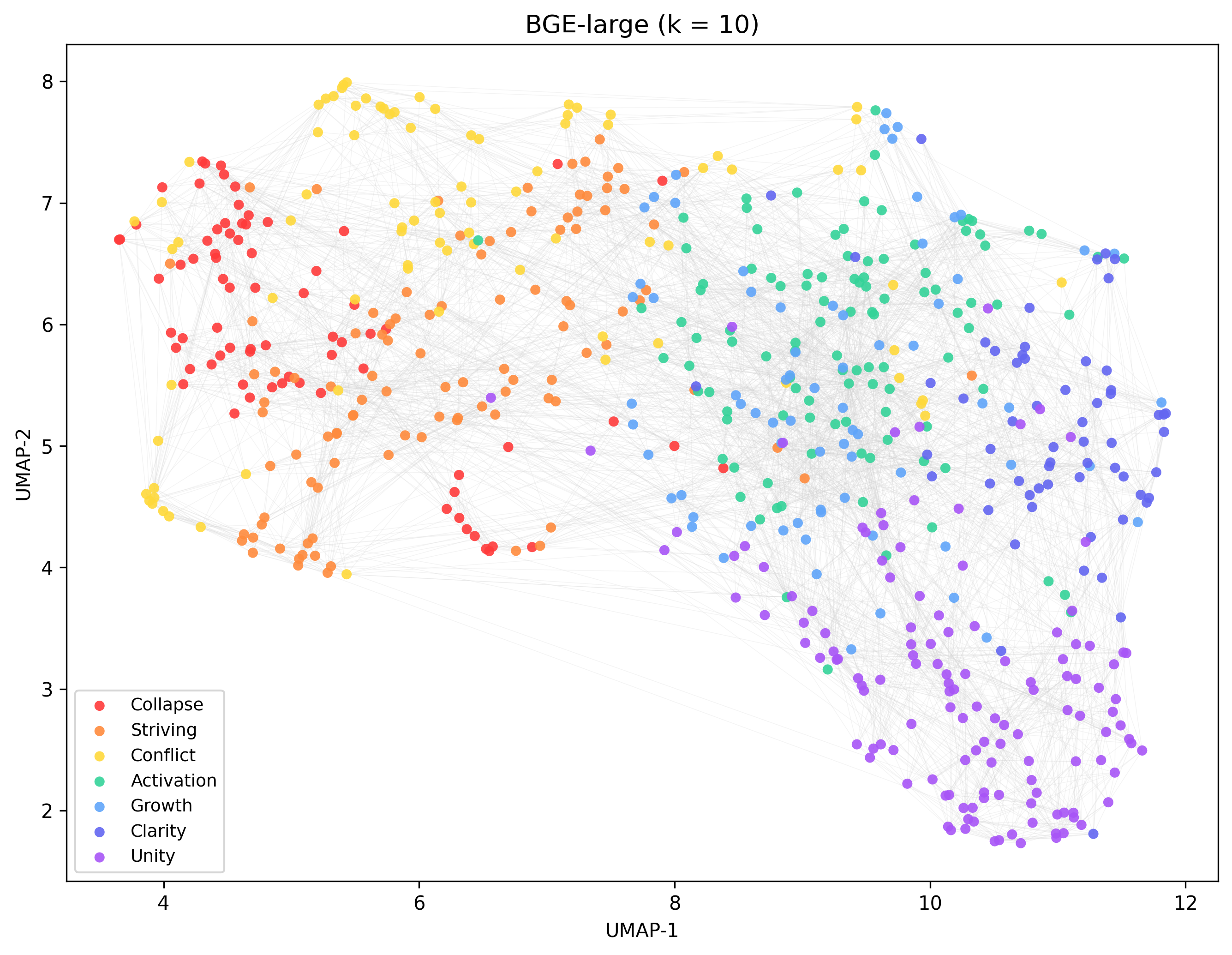}
    \caption{$k=10$}
\end{subfigure}
\hspace{0.01\textwidth}
\begin{subfigure}[b]{0.23\textwidth}
    \includegraphics[width=\linewidth]{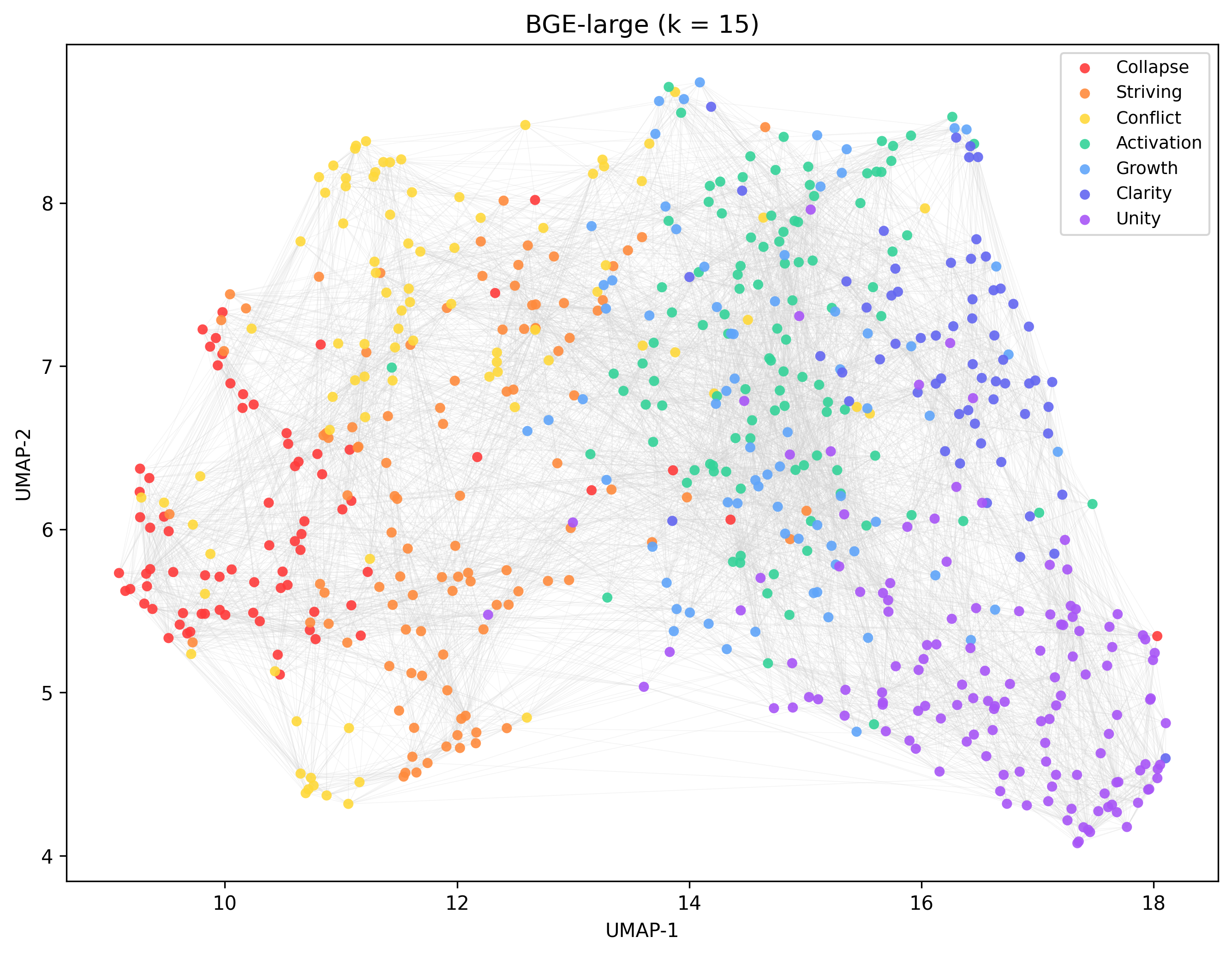}
    \caption{$k=15$}
\end{subfigure}
\hspace{0.01\textwidth}
\begin{subfigure}[b]{0.23\textwidth}
    \includegraphics[width=\linewidth]{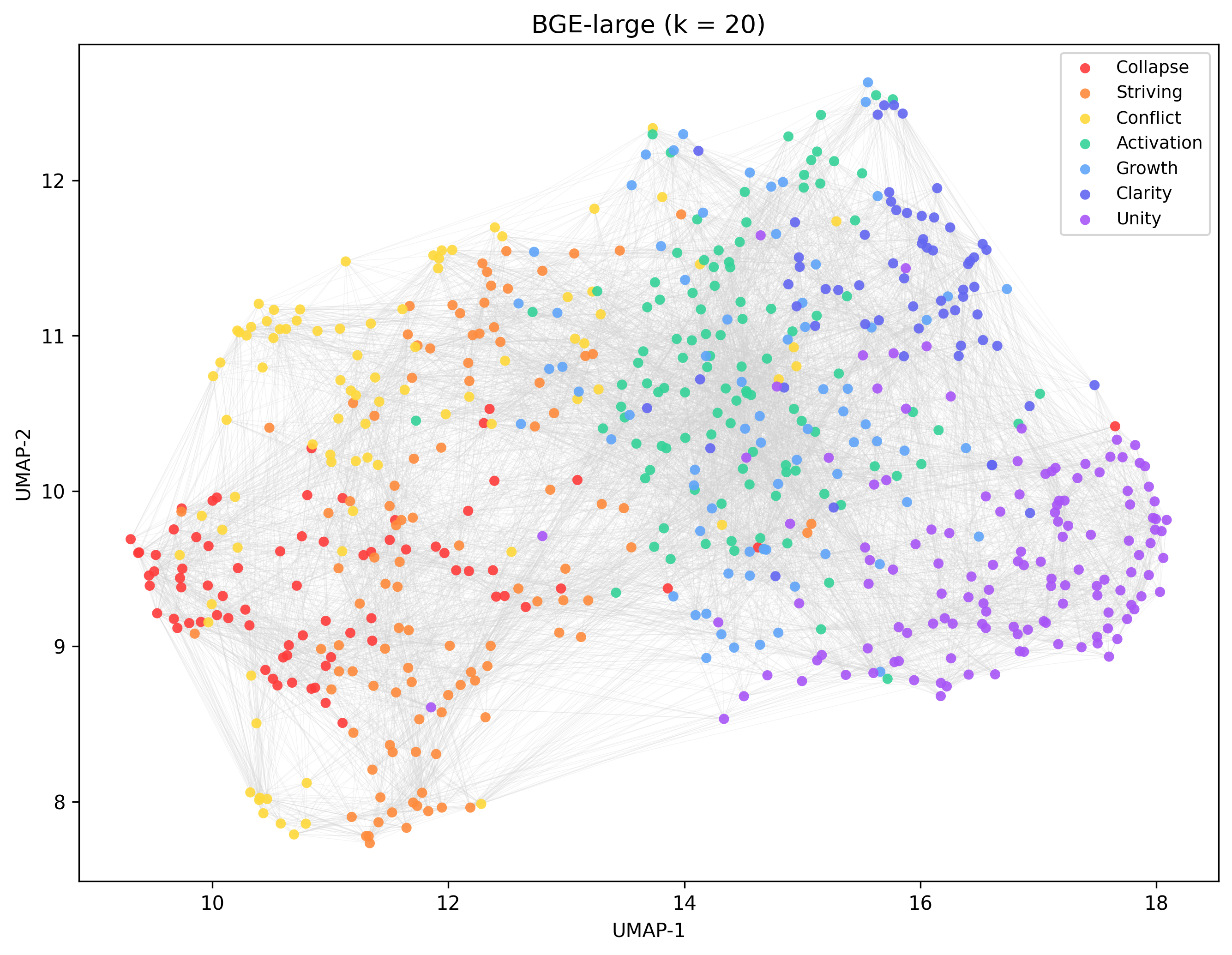}
    \caption{$k=20$}
\end{subfigure}
\hspace{0.01\textwidth}
\begin{subfigure}[b]{0.23\textwidth}
    \includegraphics[width=\linewidth]{figures/umap/umap_bge_k30.png}
    \caption{$k=30$}
\end{subfigure}

\vspace{0.2em}
{\small \textbf{BGE embeddings}}

\vspace{0.7em}

\begin{subfigure}[b]{0.23\textwidth}
    \includegraphics[width=\linewidth]{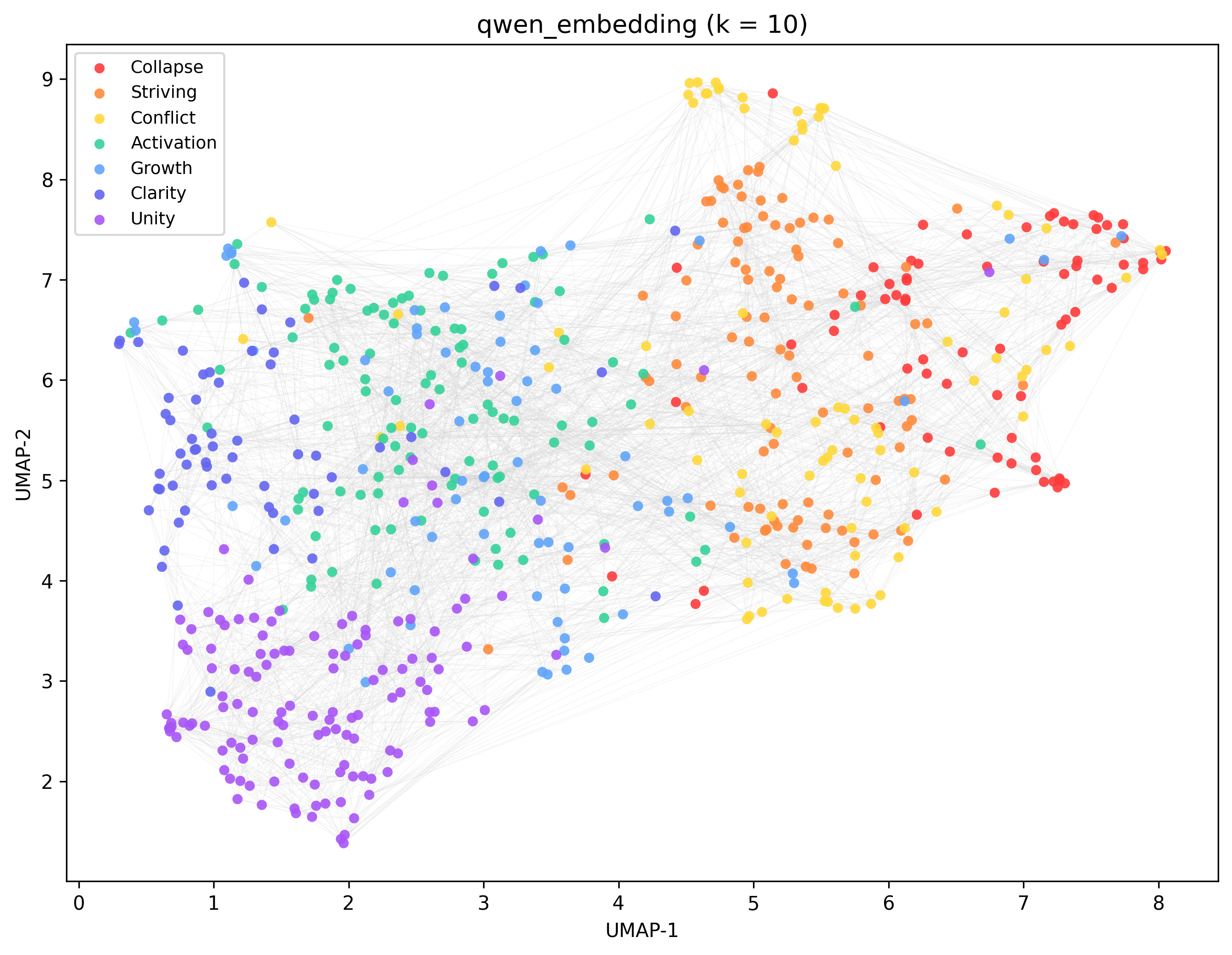}
    \caption{$k=10$}
\end{subfigure}
\hspace{0.01\textwidth}
\begin{subfigure}[b]{0.23\textwidth}
    \includegraphics[width=\linewidth]{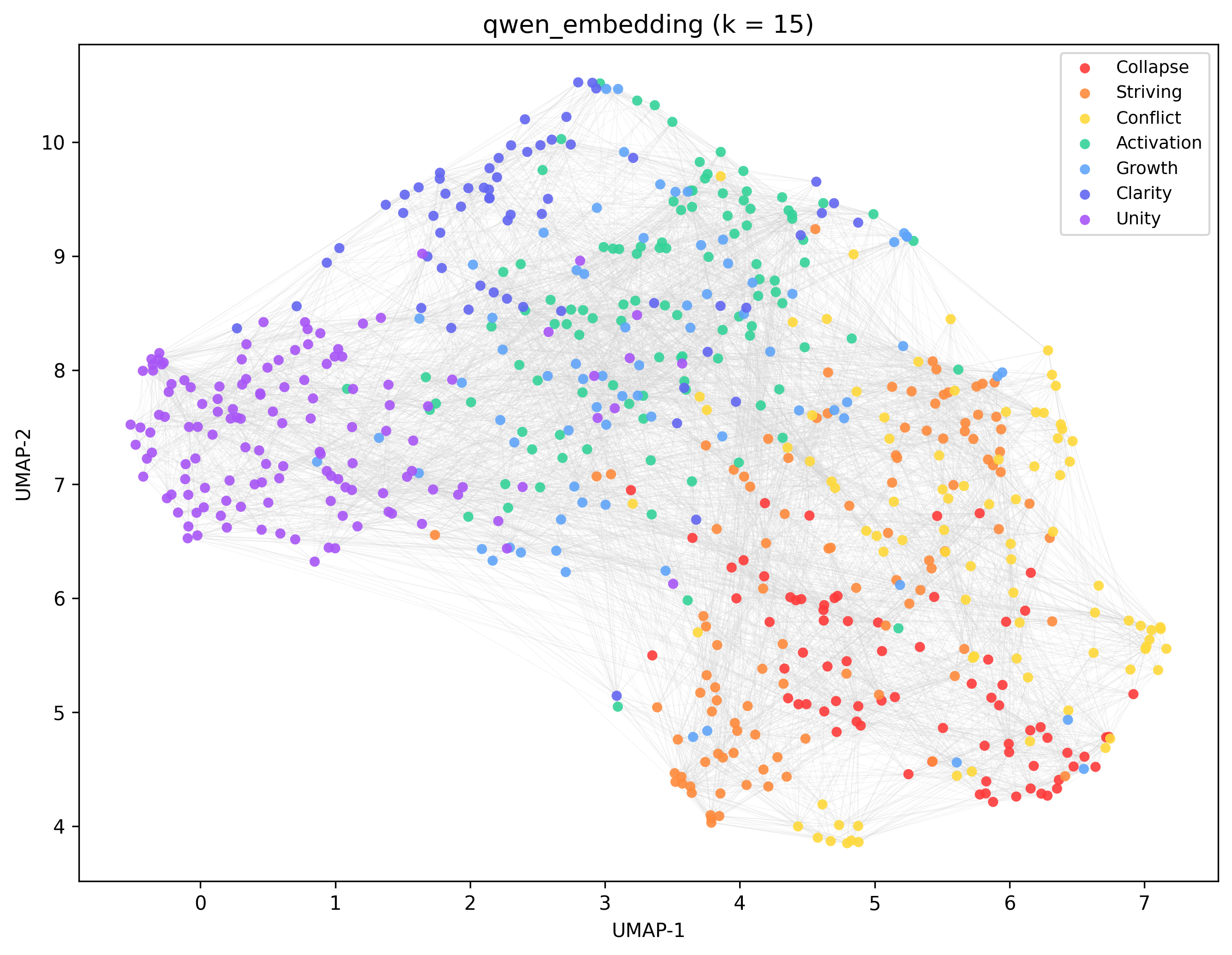}
    \caption{$k=15$}
\end{subfigure}
\hspace{0.01\textwidth}
\begin{subfigure}[b]{0.23\textwidth}
    \includegraphics[width=\linewidth]{figures/umap/umap_qwen_embedding_k20.png}
    \caption{$k=20$}
\end{subfigure}
\hspace{0.01\textwidth}
\begin{subfigure}[b]{0.23\textwidth}
    \includegraphics[width=\linewidth]{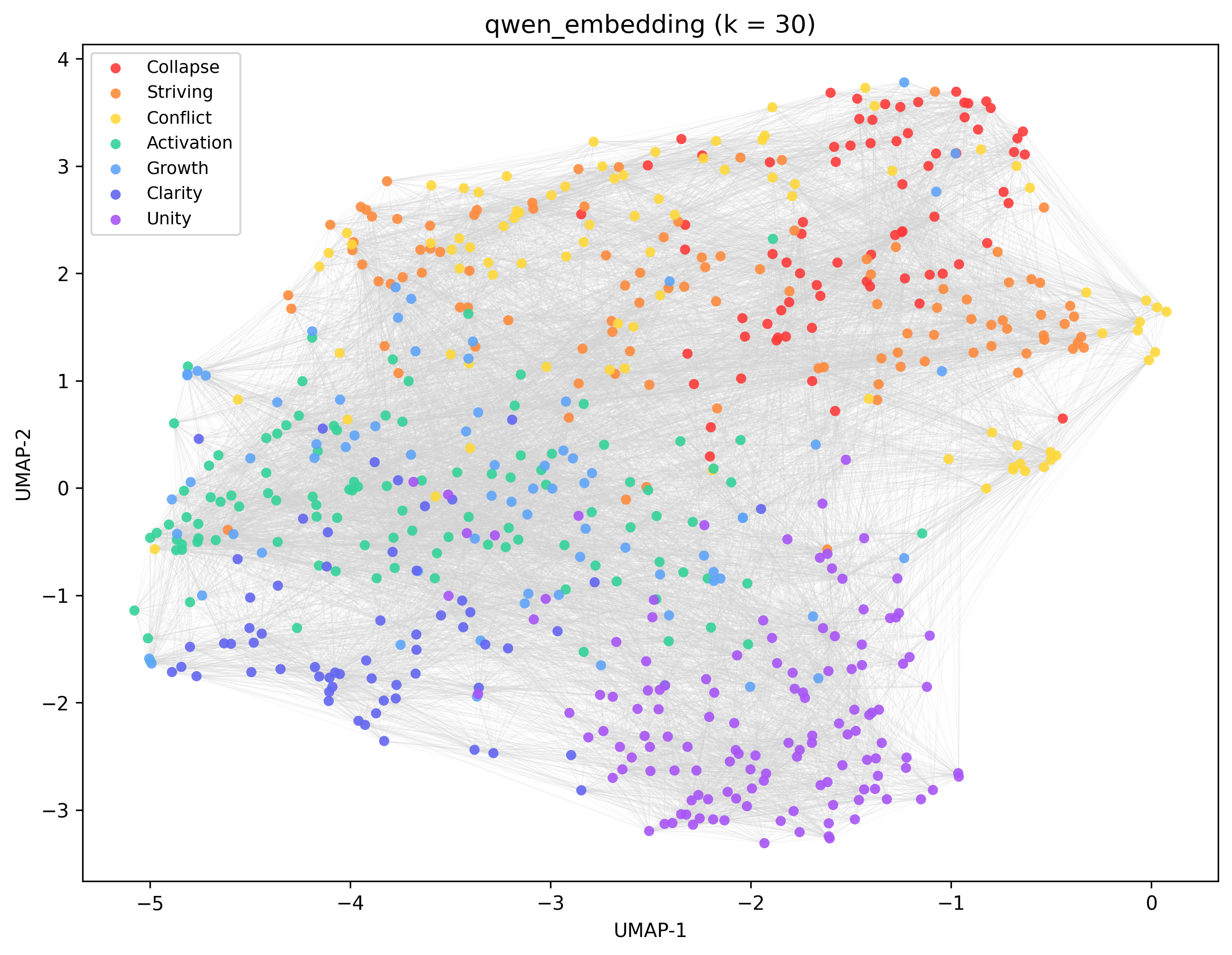}
    \caption{$k=30$}
\end{subfigure}

\vspace{0.2em}
{\small \textbf{Qwen embeddings}}

\vspace{0.7em}

\begin{subfigure}[b]{0.23\textwidth}
    \includegraphics[width=\linewidth]{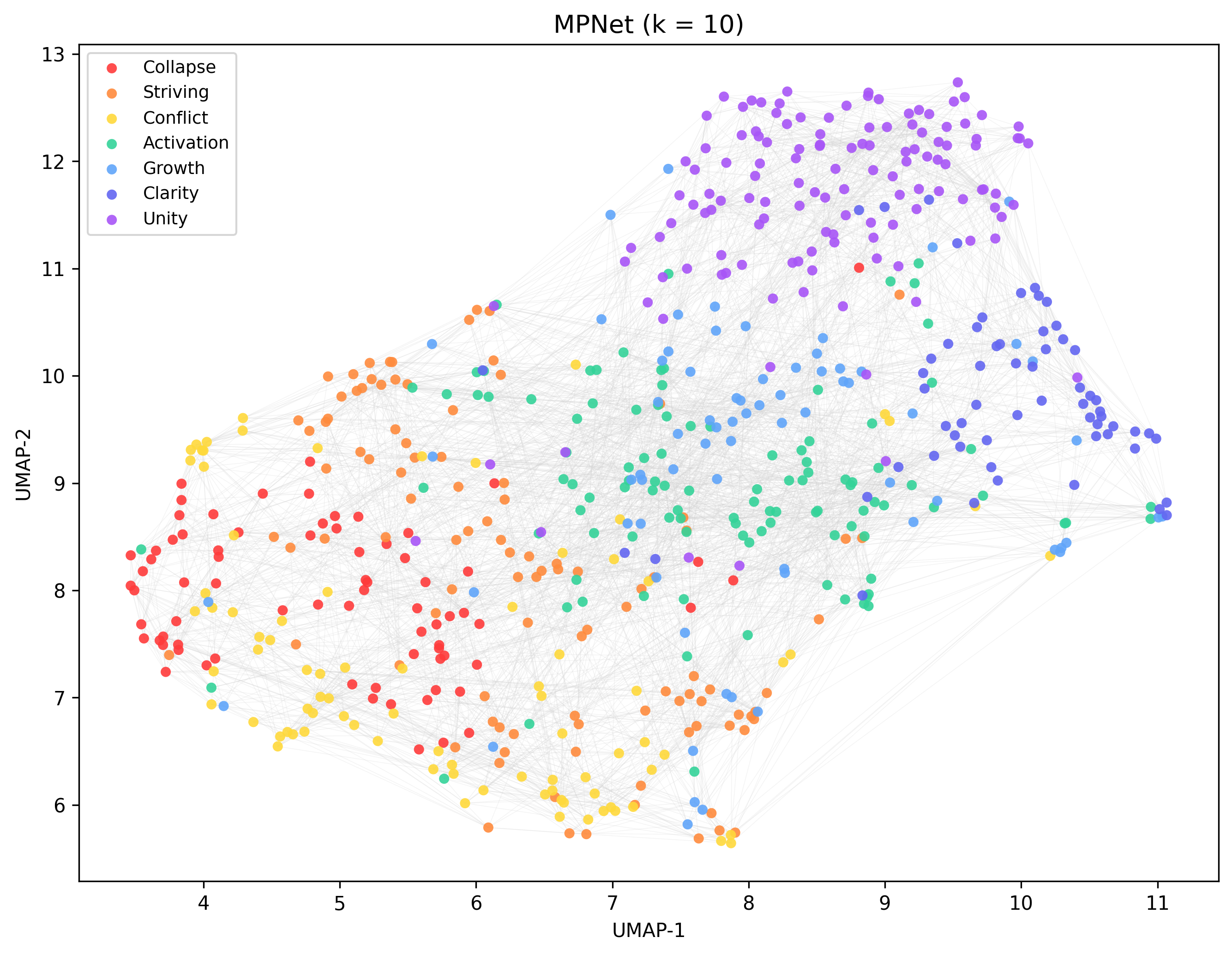}
    \caption{$k=10$}
\end{subfigure}
\hspace{0.01\textwidth}
\begin{subfigure}[b]{0.23\textwidth}
    \includegraphics[width=\linewidth]{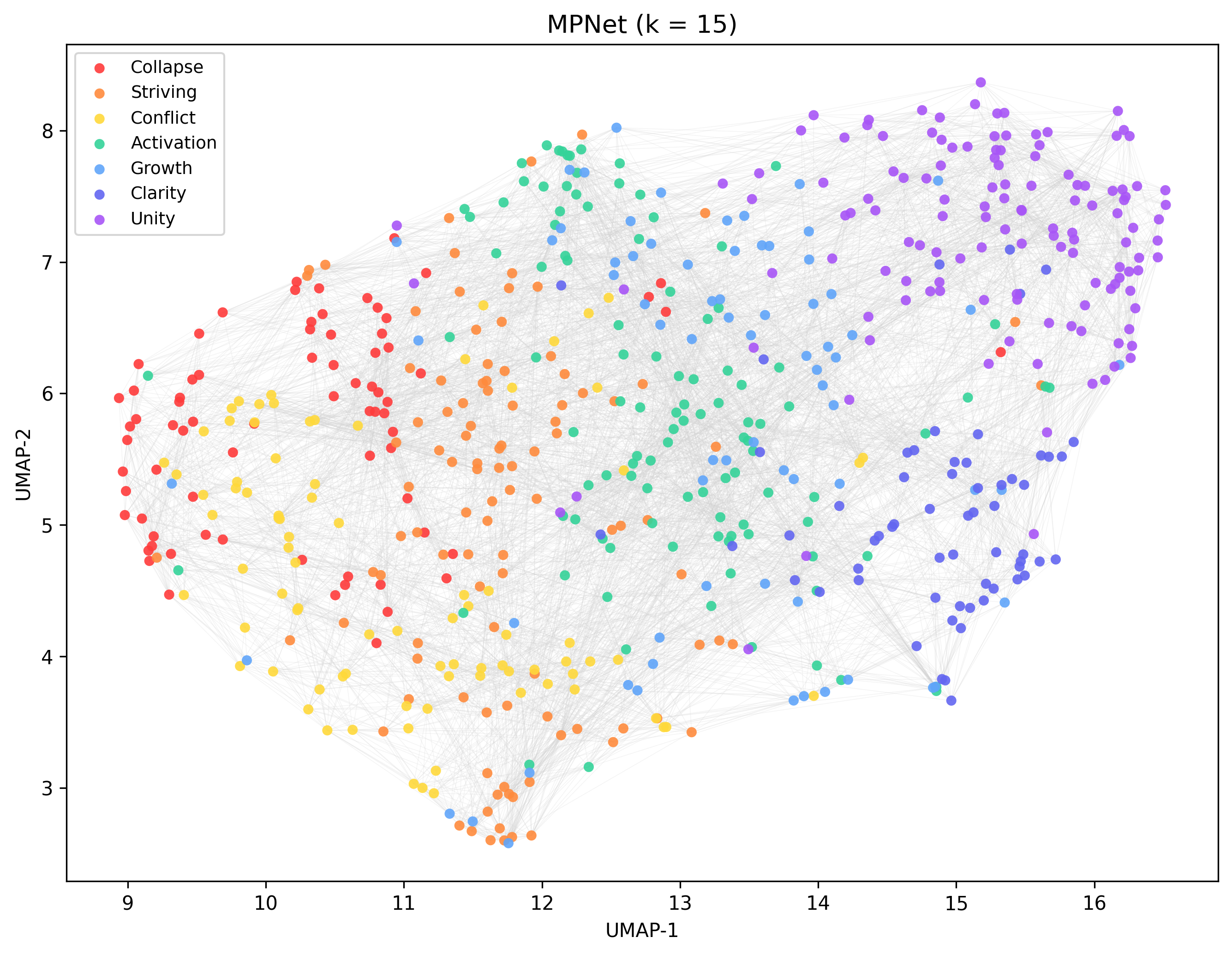}
    \caption{$k=15$}
\end{subfigure}
\hspace{0.01\textwidth}
\begin{subfigure}[b]{0.23\textwidth}
    \includegraphics[width=\linewidth]{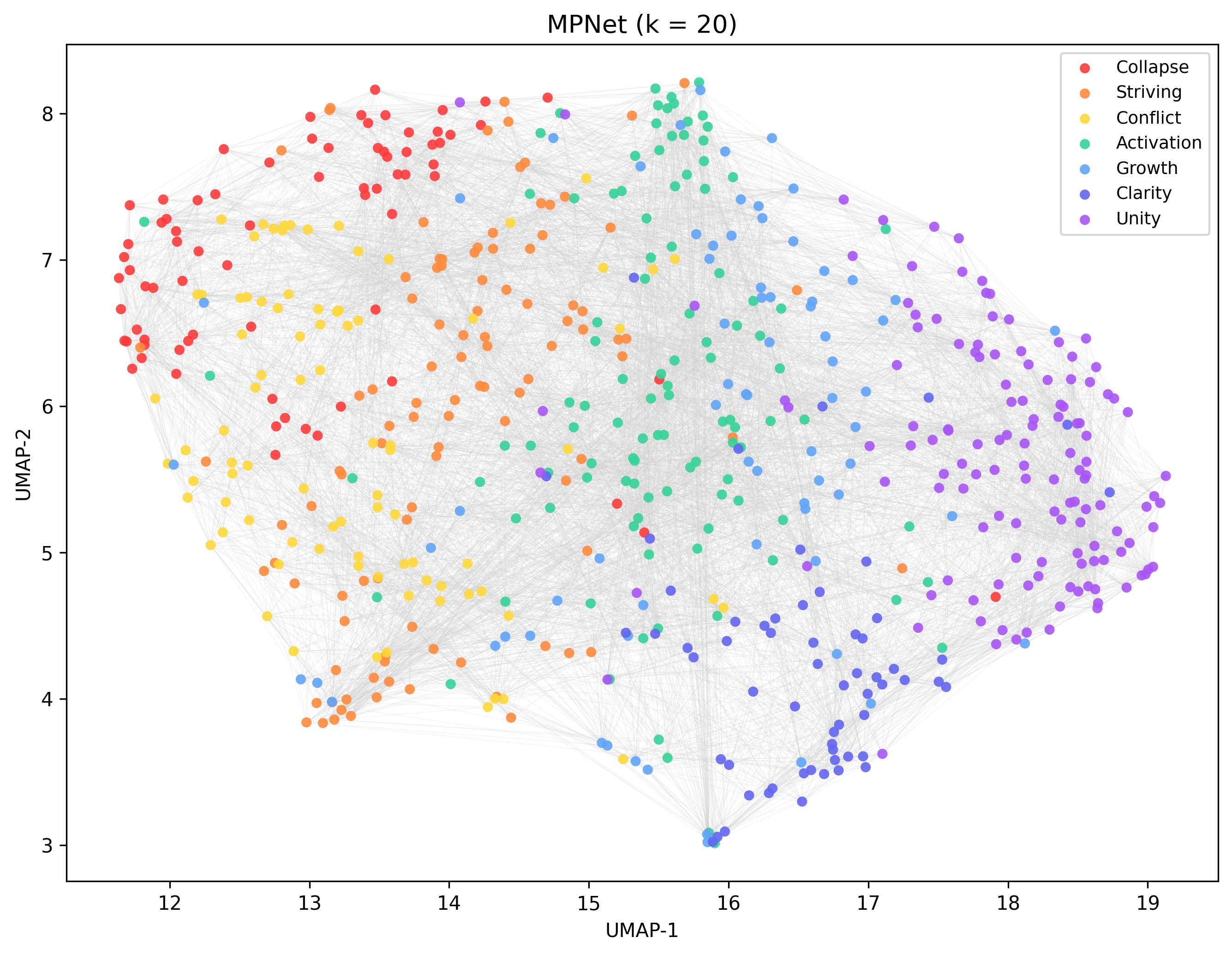}
    \caption{$k=20$}
\end{subfigure}
\hspace{0.01\textwidth}
\begin{subfigure}[b]{0.23\textwidth}
    \includegraphics[width=\linewidth]{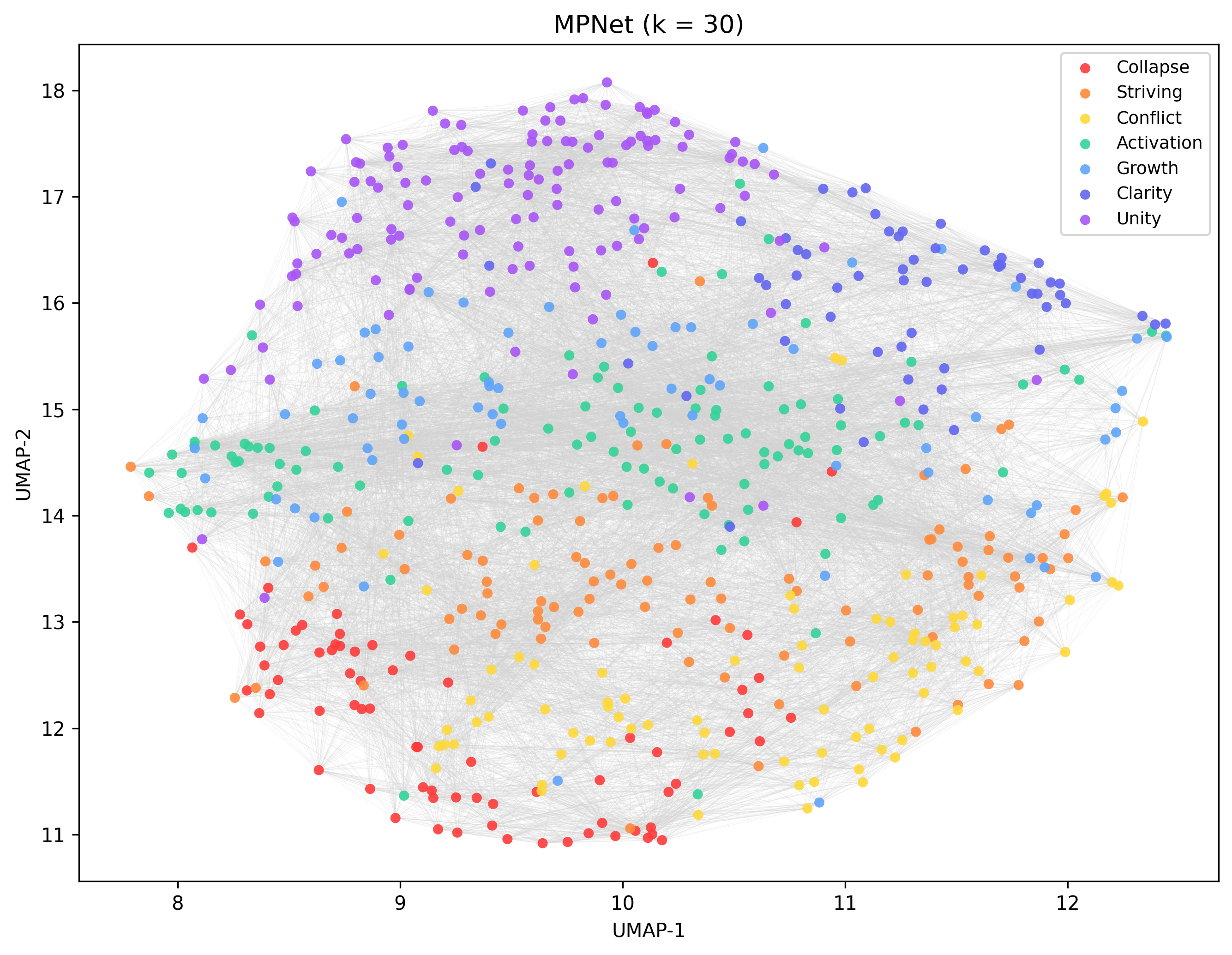}
    \caption{$k=30$}
\end{subfigure}

\vspace{0.2em}
{\small \textbf{MPNET embeddings}}

\vspace{0.7em}

\begin{subfigure}[b]{0.23\textwidth}
    \includegraphics[width=\linewidth]{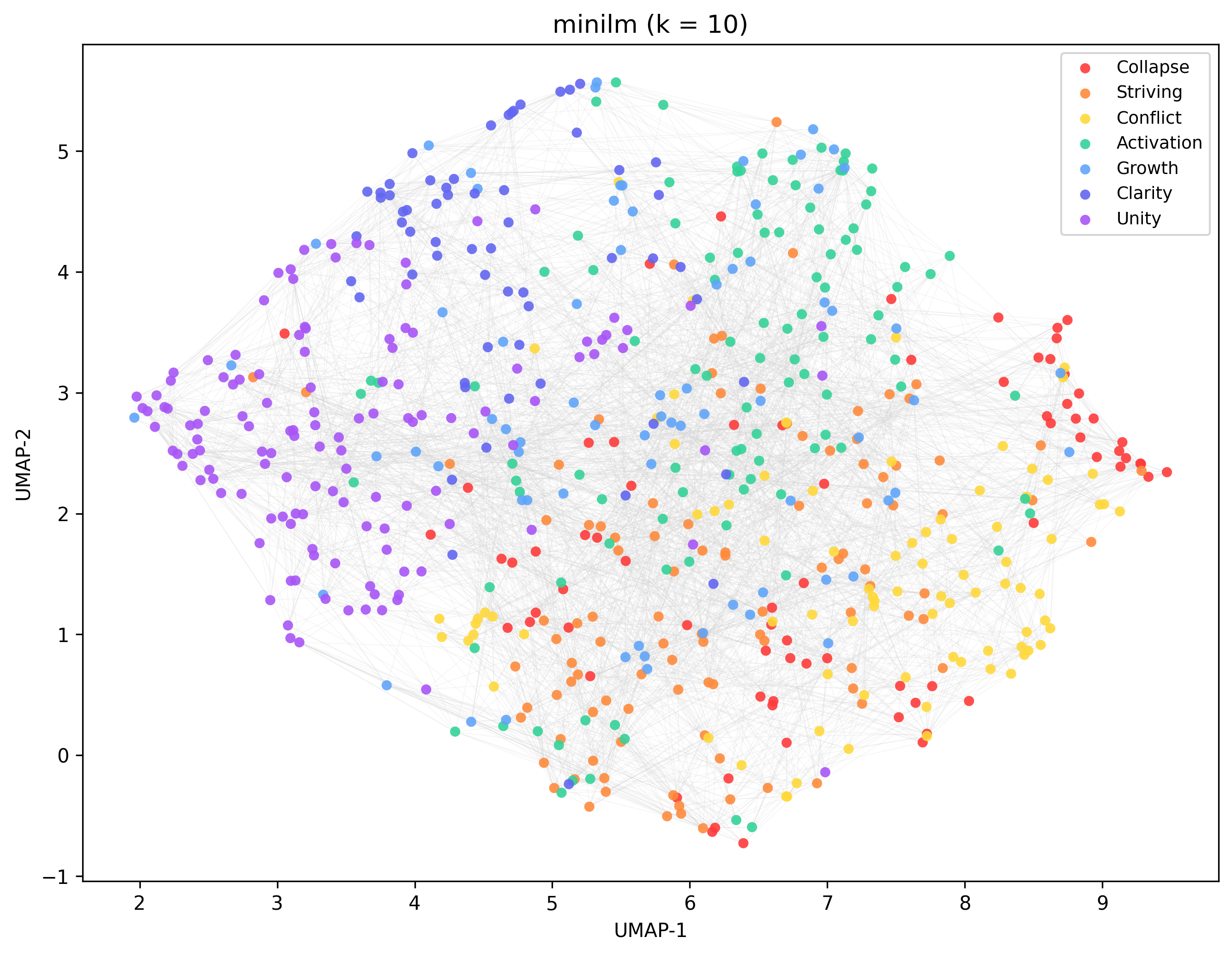}
    \caption{$k=10$}
\end{subfigure}
\hspace{0.01\textwidth}
\begin{subfigure}[b]{0.23\textwidth}
    \includegraphics[width=\linewidth]{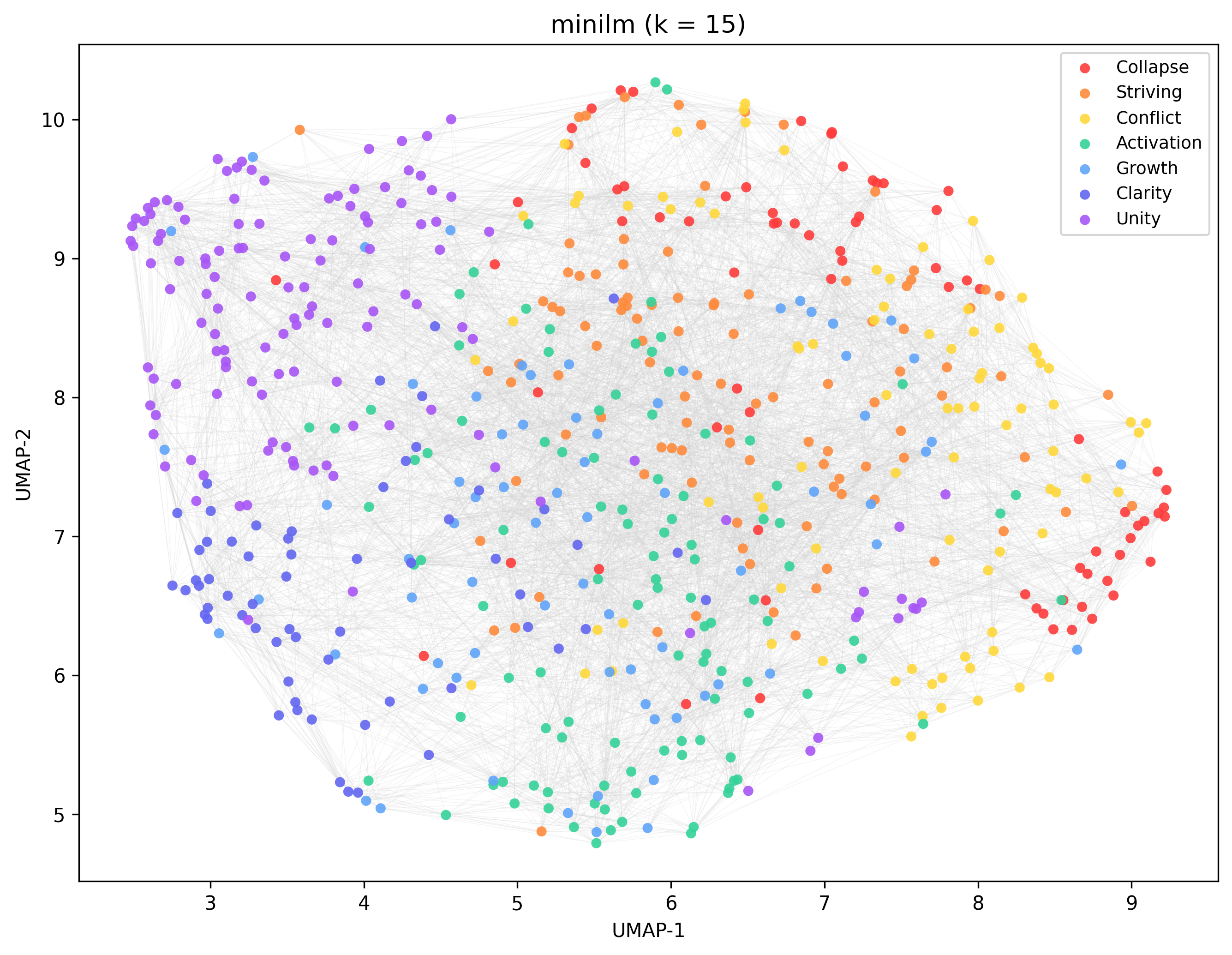}
    \caption{$k=15$}
\end{subfigure}
\hspace{0.01\textwidth}
\begin{subfigure}[b]{0.23\textwidth}
    \includegraphics[width=\linewidth]{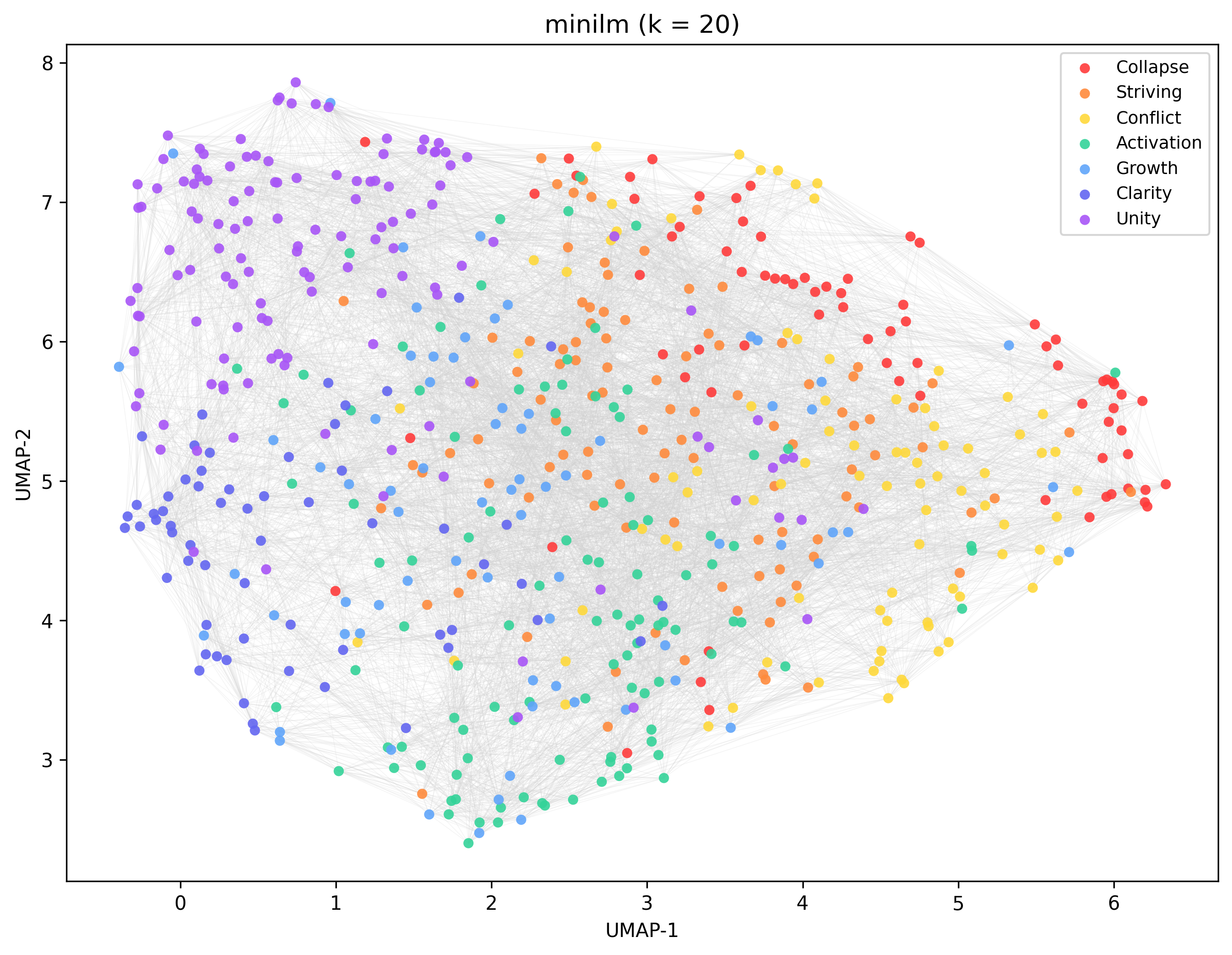}
    \caption{$k=20$}
\end{subfigure}
\hspace{0.01\textwidth}
\begin{subfigure}[b]{0.23\textwidth}
    \includegraphics[width=\linewidth]{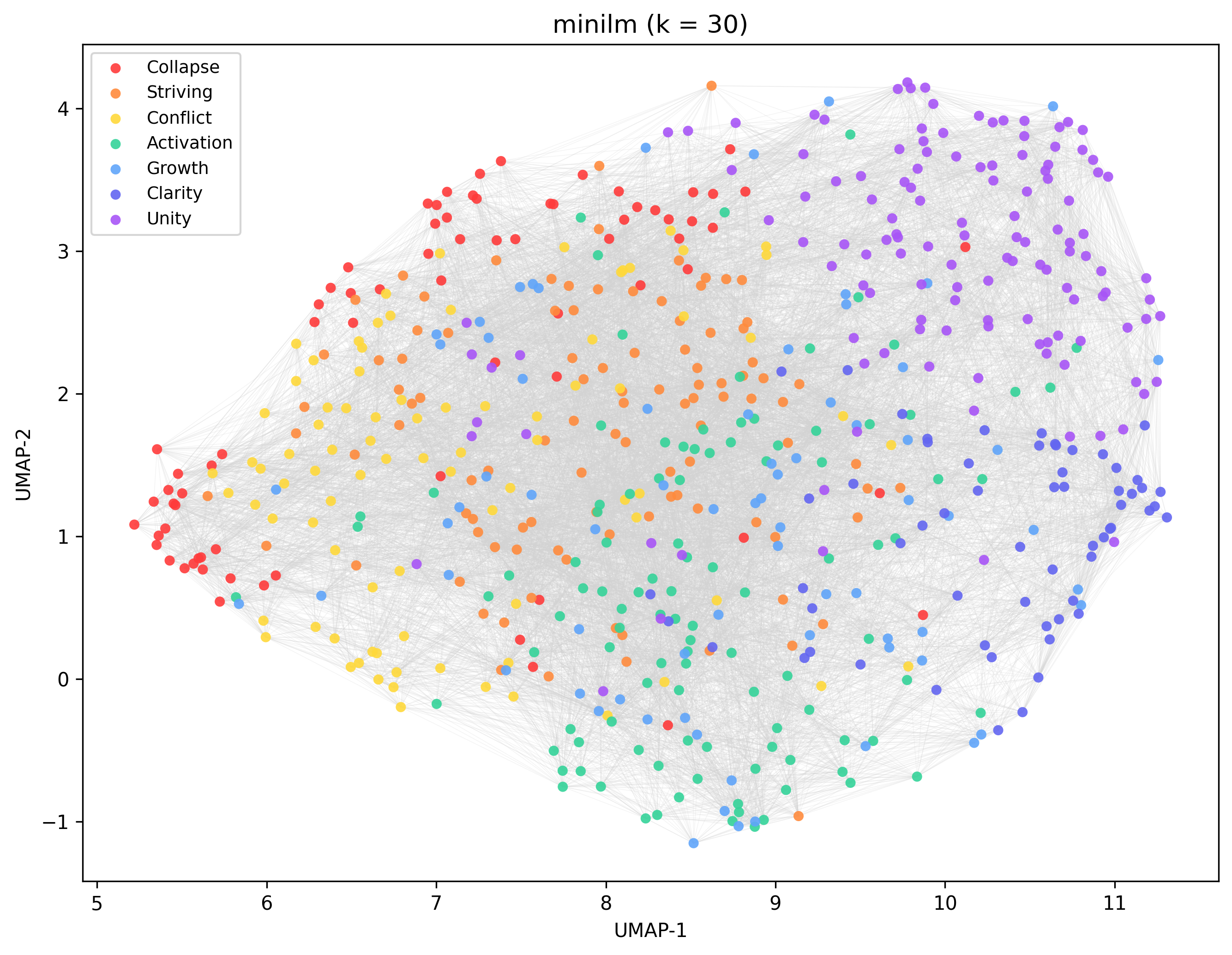}
    \caption{$k=30$}
\end{subfigure}

\vspace{0.2em}
{\small \textbf{MiniLM embeddings}}

\vspace{-0.3em}

\caption{\small
UMAP projections across neighborhood scales $k \in \{10,15,20,30\}$ for BGE, Qwen, MPNET, and MiniLM embeddings. Across models and scales, the overall tier organization remains stable, while larger neighborhood sizes produce smoother local structure, suggesting that the observed manifold geometry is robust to the choice of connectivity scale.
}
\label{fig:all_umap_models}

\vspace{-0.8em}
\end{figure*}

Figure~\ref{fig:all_umap_models} presents UMAP visualizations across all embedding models and neighborhood sizes $k$.

\FloatBarrier
\section{Full Overall Tier Transition Smoothness (TTS) results}
\label{app: full TTS}

\begin{table}[htbp!]
\centering
\scriptsize
\caption{Overall Tier Transition Smoothness (TTS) across embedding models and neighborhood sizes. Values report the proportion of graph edges with ordinal jump magnitude $0$, $1$, $2$, or greater than $2$. Bold values indicate the best result within each $k$ (higher is better for jump0 and jump1; lower is better for jump2 and jump$>2$).}
\label{tab:tts_overall}
\begin{tabular}{llcccc}
\toprule
Model & $k$ & jump0 & jump1 & jump2 & jump$>2$ \\
\midrule
BGE    & 10 & 0.539 & \textbf{0.226} & 0.144 & \textbf{0.091} \\
MPNet  & 10 & 0.523 & 0.207 & 0.155 & 0.114 \\
MiniLM & 10 & 0.444 & 0.217 & 0.170 & 0.169 \\
Qwen   & 10 & \textbf{0.551} & 0.216 & \textbf{0.131} & 0.102 \\
\midrule
BGE    & 15 & 0.508 & \textbf{0.239} & 0.154 & \textbf{0.099} \\
MPNet  & 15 & 0.481 & 0.222 & 0.163 & 0.133 \\
MiniLM & 15 & 0.415 & 0.226 & 0.176 & 0.183 \\
Qwen   & 15 & \textbf{0.513} & 0.227 & \textbf{0.143} & 0.117 \\
\midrule
BGE    & 20 & 0.479 & \textbf{0.250} & 0.161 & \textbf{0.109} \\
MPNet  & 20 & 0.462 & 0.227 & 0.167 & 0.143 \\
MiniLM & 20 & 0.396 & 0.231 & 0.181 & 0.193 \\
Qwen   & 20 & \textbf{0.485} & 0.235 & \textbf{0.152} & 0.128 \\
\midrule
BGE    & 30 & 0.444 & \textbf{0.260} & 0.169 & \textbf{0.127} \\
MPNet  & 30 & 0.423 & 0.237 & 0.176 & 0.163 \\
MiniLM & 30 & 0.367 & 0.239 & 0.188 & 0.182 \\
Qwen   & 30 & \textbf{0.449} & 0.248 & \textbf{0.165} & 0.138 \\
\bottomrule
\end{tabular}
\end{table}

\FloatBarrier
\section{Full Geodesic--Euclidean Stretch (GES) Results}
\label{app: full GES}

\begin{table}[htbp!]
\centering
\scriptsize
\caption{Overall Geodesic-Euclidean Stretch (GES) across models and neighborhood sizes. Values greater than 1 indicate that intrinsic graph-based distances exceed ambient Euclidean distances.}
\begin{tabular}{llccc}
\toprule
Model & $k$ & Mean & Median & Std \\
\midrule
BGE    & 10 & 2.355 & 2.347 & 0.491 \\
BGE    & 15 & 2.119 & 2.218 & 0.419 \\
BGE    & 20 & 1.974 & 2.108 & 0.383 \\
BGE    & 30 & 1.821 & 1.723 & 0.349 \\
\midrule
MPNet  & 10 & 2.407 & 2.412 & 0.490 \\
MPNet  & 15 & 2.155 & 2.291 & 0.423 \\
MPNet  & 20 & 2.015 & 2.155 & 0.396 \\
MPNet  & 30 & 1.844 & 1.745 & 0.361 \\
\midrule
MiniLM & 10 & 2.387 & 2.435 & 0.476 \\
MiniLM & 15 & 2.160 & 2.332 & 0.414 \\
MiniLM & 20 & 2.021 & 1.877 & 0.395 \\
MiniLM & 30 & 1.845 & 1.759 & 0.354 \\
\midrule
Qwen   & 10 & 2.351 & 2.361 & 0.481 \\
Qwen   & 15 & 2.117 & 2.244 & 0.412 \\
Qwen   & 20 & 1.987 & 2.111 & 0.385 \\
Qwen   & 30 & 1.819 & 1.726 & 0.356 \\
\bottomrule
\end{tabular}
\end{table}

\FloatBarrier
\section{Full GES Heatmaps}
\label{app:full_ges_heatmaps}

\begin{figure*}[htbp!]
\centering
\begin{subfigure}[b]{0.23\textwidth}
\includegraphics[width=\linewidth]{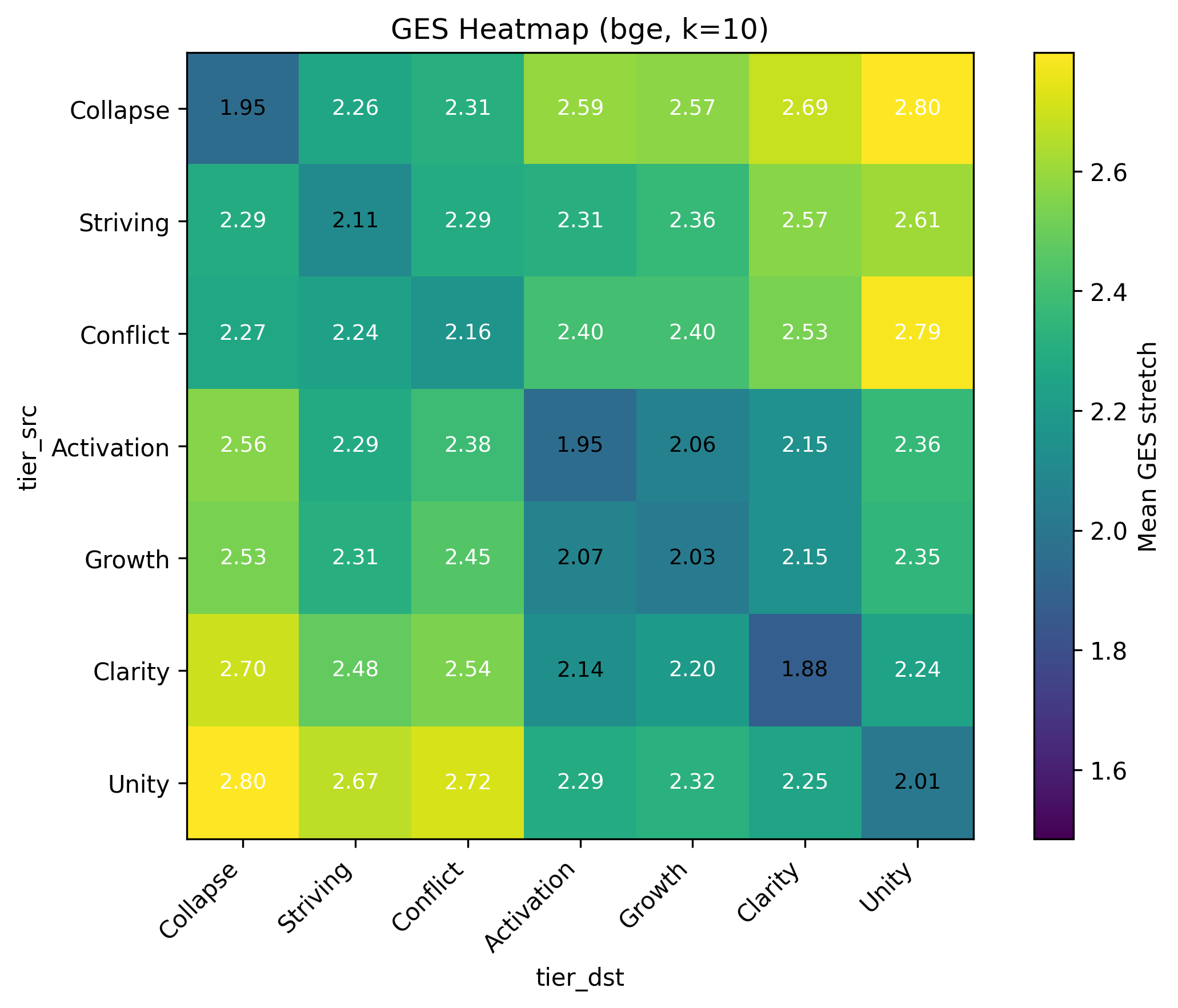}
\caption{$k=10$}
\end{subfigure}
\hspace{0.01\textwidth}
\begin{subfigure}[b]{0.23\textwidth}
\includegraphics[width=\linewidth]{figures/ges/ges_heatmap_bge_k15.png}
\caption{$k=15$}
\end{subfigure}
\hspace{0.01\textwidth}
\begin{subfigure}[b]{0.23\textwidth}
\includegraphics[width=\linewidth]{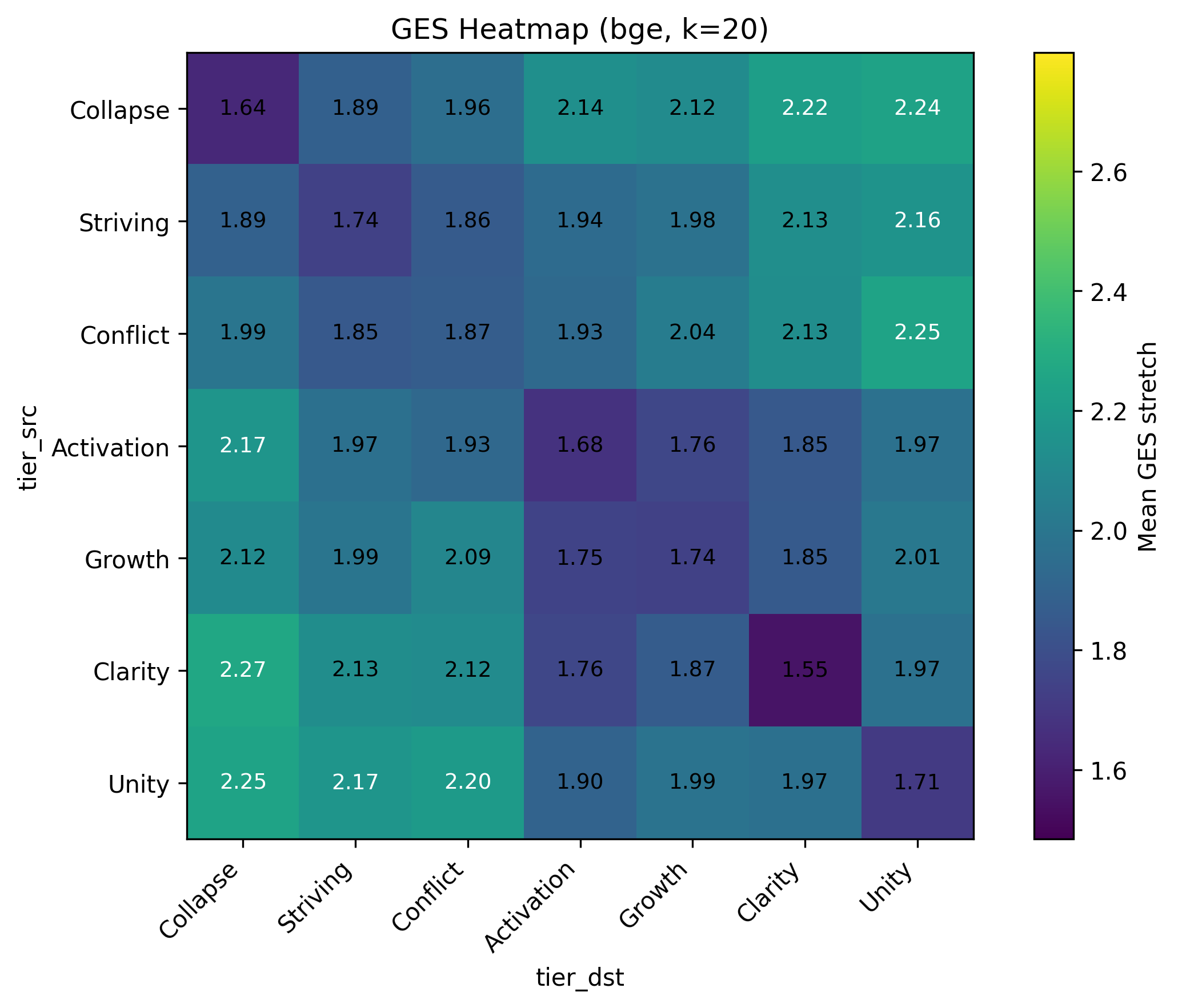}
\caption{$k=20$}
\end{subfigure}
\hspace{0.01\textwidth}
\begin{subfigure}[b]{0.23\textwidth}
\includegraphics[width=\linewidth]{figures/ges/ges_heatmap_bge_k30.png}
\caption{$k=30$}
\end{subfigure}

\vspace{-0.3em}
\caption{\small Tier-pair Geodesic--Euclidean Stretch (GES) for BGE embeddings across $k \in \{10,15,20,30\}$. Within-tier pairs show lower stretch, while distant tier pairs show higher values. Absolute stretch decreases as $k$ increases, but the overall structure remains stable.}
\label{fig:ges_bge_heatmaps}
\vspace{-0.8em}
\end{figure*}

\FloatBarrier
\section{Full Convexity Results}
\label{app:full_convexity}

\begin{table*}[htbp!]
\centering
\scriptsize
\caption{Mean geodesic containment by tier across embedding models and neighborhood sizes. Higher values indicate that shortest paths between within-tier node pairs remain more strongly inside the same tier region. Values greater than 0.9 are highlighted in bold.}
\label{tab:convexity_allk}
\begin{tabular}{llccccccc}
\toprule
Model & $k$ & Collapse & Striving & Conflict & Activation & Growth & Clarity & Unity \\
\midrule
BGE & 10 & \textbf{0.934} & 0.871 & 0.758 & 0.888 & 0.710 & 0.879 & \textbf{0.917} \\
BGE & 15 & \textbf{0.948} & 0.882 & 0.779 & \textbf{0.906} & 0.736 & 0.894 & \textbf{0.928} \\
BGE & 20 & \textbf{0.944} & 0.899 & 0.796 & \textbf{0.913} & 0.746 & \textbf{0.905} & \textbf{0.934} \\
BGE & 30 & \textbf{0.948} & \textbf{0.909} & 0.830 & \textbf{0.924} & 0.778 & \textbf{0.914} & \textbf{0.946} \\
\midrule
MPNet & 10 & \textbf{0.914} & 0.824 & 0.795 & 0.818 & 0.746 & 0.900 & \textbf{0.951} \\
MPNet & 15 & \textbf{0.934} & 0.836 & 0.812 & 0.840 & 0.773 & \textbf{0.901} & \textbf{0.954} \\
MPNet & 20 & \textbf{0.940} & 0.857 & 0.834 & 0.855 & 0.791 & \textbf{0.913} & \textbf{0.959} \\
MPNet & 30 & \textbf{0.947} & 0.872 & 0.861 & 0.877 & 0.813 & \textbf{0.923} & \textbf{0.960} \\
\midrule
MiniLM & 10 & \textbf{0.904} & 0.814 & 0.731 & 0.793 & 0.695 & 0.862 & \textbf{0.904} \\
MiniLM & 15 & \textbf{0.916} & 0.839 & 0.771 & 0.811 & 0.730 & 0.889 & \textbf{0.917} \\
MiniLM & 20 & \textbf{0.928} & 0.855 & 0.790 & 0.837 & 0.741 & 0.899 & \textbf{0.928} \\
MiniLM & 30 & \textbf{0.934} & 0.869 & 0.817 & 0.859 & 0.779 & \textbf{0.915} & \textbf{0.934} \\
\midrule
Qwen & 10 & \textbf{0.941} & 0.888 & 0.806 & 0.840 & 0.715 & 0.874 & \textbf{0.961} \\
Qwen & 15 & \textbf{0.945} & 0.899 & 0.819 & 0.854 & 0.744 & 0.896 & \textbf{0.964} \\
Qwen & 20 & \textbf{0.944} & \textbf{0.907} & 0.837 & 0.871 & 0.774 & \textbf{0.907} & \textbf{0.966} \\
Qwen & 30 & \textbf{0.954} & \textbf{0.910} & 0.868 & 0.882 & 0.799 & \textbf{0.934} & \textbf{0.968} \\
\bottomrule
\end{tabular}
\end{table*}

\begin{table*}[htbp]
\centering
\scriptsize
\caption{Full geodesic containment rate across embedding models and neighborhood sizes. Values indicate the proportion of within-tier shortest paths that remain entirely inside the same tier. Values greater than 0.7 are highlighted in bold.}
\label{tab:convexity_full_all}
\begin{tabular}{llccccccc}
\toprule
Model & $k$ & Collapse & Striving & Conflict & Activation & Growth & Clarity & Unity \\
\midrule
BGE & 10 & \textbf{0.804} & 0.646 & 0.400 & 0.652 & 0.248 & 0.649 & \textbf{0.772} \\
BGE & 15 & \textbf{0.844} & 0.660 & 0.432 & \textbf{0.711} & 0.267 & 0.694 & \textbf{0.798} \\
BGE & 20 & \textbf{0.835} & \textbf{0.704} & 0.456 & \textbf{0.736} & 0.275 & \textbf{0.723} & \textbf{0.813} \\
BGE & 30 & \textbf{0.849} & \textbf{0.729} & 0.529 & \textbf{0.769} & 0.345 & \textbf{0.745} & \textbf{0.844} \\
\midrule
MPNet & 10 & \textbf{0.734} & 0.496 & 0.467 & 0.468 & 0.348 & \textbf{0.710} & \textbf{0.858} \\
MPNet & 15 & \textbf{0.804} & 0.526 & 0.498 & 0.531 & 0.384 & \textbf{0.710} & \textbf{0.866} \\
MPNet & 20 & \textbf{0.824} & 0.580 & 0.548 & 0.580 & 0.419 & \textbf{0.745} & \textbf{0.878} \\
MPNet & 30 & \textbf{0.843} & 0.623 & 0.611 & 0.640 & 0.457 & \textbf{0.770} & \textbf{0.879} \\
\midrule
MiniLM & 10 & 0.633 & 0.470 & 0.332 & 0.424 & 0.214 & 0.629 & \textbf{0.727} \\
MiniLM & 15 & 0.663 & 0.526 & 0.397 & 0.470 & 0.281 & 0.698 & \textbf{0.761} \\
MiniLM & 20 & 0.671 & 0.569 & 0.433 & 0.533 & 0.285 & \textbf{0.711} & \textbf{0.794} \\
MiniLM & 30 & 0.697 & 0.612 & 0.482 & 0.587 & 0.367 & \textbf{0.751} & \textbf{0.806} \\
\midrule
Qwen & 10 & \textbf{0.824} & 0.661 & 0.486 & 0.536 & 0.260 & 0.664 & \textbf{0.882} \\
Qwen & 15 & \textbf{0.838} & 0.693 & 0.507 & 0.572 & 0.299 & \textbf{0.715} & \textbf{0.889} \\
Qwen & 20 & \textbf{0.836} & \textbf{0.720} & 0.544 & 0.618 & 0.359 & \textbf{0.737} & \textbf{0.897} \\
Qwen & 30 & \textbf{0.860} & \textbf{0.729} & 0.621 & 0.649 & 0.409 & \textbf{0.806} & \textbf{0.902} \\
\bottomrule
\end{tabular}
\end{table*}

\FloatBarrier

\section{Examples of Full Trajectories}
\label{app:full_trajactories}
\paragraph{Example 1.}
\noindent
\textit{This trajectory reflects a progressive transition from collapse-related self-negation and defensive striving toward emotional regulation, cognitive coherence, and ultimately non-dual integrative awareness.
}

\begin{figure}[htbp!]
\centering

\scriptsize
\renewcommand{\arraystretch}{0.92}

\begin{tabular}{c p{0.8cm} p{0.6cm} p{8.5cm}}
\toprule
 & Tier & Score & Sentence \\
\midrule

$\triangleright$ 
& \cellcolor{CollapseColor} Collapse 
& \cellcolor{CollapseColor} -4.80 
& I feel worthless no matter what I do. \\

& \cellcolor{CollapseColor} Collapse 
& \cellcolor{CollapseColor} -4.78 
& I feel like everyone hates me. \\

& \cellcolor{CollapseColor} Collapse 
& \cellcolor{CollapseColor} -4.55 
& I feel like I deserve punishment. \\

& \cellcolor{CollapseColor} Collapse 
& \cellcolor{CollapseColor} -4.50 
& A part of me wants vengeance, even though it hurts. \\

& \cellcolor{CollapseColor} Collapse 
& \cellcolor{CollapseColor} -4.56 
& Even when it's not my fault, I feel to blame. \\

& \cellcolor{StrivingColor} Striving 
& \cellcolor{StrivingColor} -2.70 
& I call it leadership, but really I just can’t stand feeling ignored. \\

& \cellcolor{StrivingColor} Striving 
& \cellcolor{StrivingColor} -1.83 
& Rest feels undeserved, so I keep going. \\

& \cellcolor{StrivingColor} Striving 
& \cellcolor{StrivingColor} -2.94 
& I chase anything that keeps me from sitting with myself. \\

& \cellcolor{ConflictColor} Conflict 
& \cellcolor{ConflictColor} -1.60 
& That scoff keeps them guessing while I steady myself. \\

& \cellcolor{StrivingColor} Striving 
& \cellcolor{StrivingColor} -1.82 
& I try to stay ahead so I won’t be judged. \\

& \cellcolor{StrivingColor} Striving 
& \cellcolor{StrivingColor} -2.94 
& I keep myself busy so I don’t feel the heaviness underneath. \\

& \cellcolor{ActivationColor} Activation 
& \cellcolor{ActivationColor} 0.50 
& I’m trying to let things be long enough for me to regain my footing. \\

& \cellcolor{ActivationColor} Activation 
& \cellcolor{ActivationColor} 0.50 
& I’m loosening my grip just enough to let things move again. \\

& \cellcolor{ActivationColor} Activation 
& \cellcolor{ActivationColor} 0.45 
& I loosen my emotional grip so I can act clearly. \\

& \cellcolor{GrowthColor} Growth 
& \cellcolor{GrowthColor} 1.70 
& I’m learning to stay more relaxed when work is stressful. \\

& \cellcolor{GrowthColor} Growth 
& \cellcolor{GrowthColor} 1.70 
& No matter how busy I am, I take breaks to relax my eyes and body. \\

& \cellcolor{ClarityColor} Clarity 
& \cellcolor{ClarityColor} 2.65 
& I’m able to adjust my view without feeling like I’m losing ground. \\

& \cellcolor{GrowthColor} Growth 
& \cellcolor{GrowthColor} 0.55 
& It helps me grow when I see things honestly. \\

& \cellcolor{GrowthColor} Growth 
& \cellcolor{GrowthColor} 1.20 
& It feels good to be part of something that helps others grow. \\

& \cellcolor{GrowthColor} Growth 
& \cellcolor{GrowthColor} 1.10 
& Progress feels steady and real. \\

& \cellcolor{ClarityColor} Clarity 
& \cellcolor{ClarityColor} 2.69 
& Understanding feels complete enough to act, even without absolute certainty. \\

& \cellcolor{ActivationColor} Activation 
& \cellcolor{ActivationColor} 0.04 
& Existence itself is not conditional. \\

& \cellcolor{UnityColor} Unity 
& \cellcolor{UnityColor} 4.60 
& In seeing, there is no seer, only awareness aware. \\

& \cellcolor{UnityColor} Unity 
& \cellcolor{UnityColor} 4.55 
& There is no meditator, only meditation. \\

& \cellcolor{UnityColor} Unity 
& \cellcolor{UnityColor} 3.90 
& Joy doesn't need a reason, it's the soul remembering itself. \\

$\triangleleft$ 
& \cellcolor{UnityColor} Unity 
& \cellcolor{UnityColor} 4.15 
& Bliss is silence smiling. \\

\end{tabular}

\vspace{-0.2em}

\includegraphics[width=0.62\linewidth]{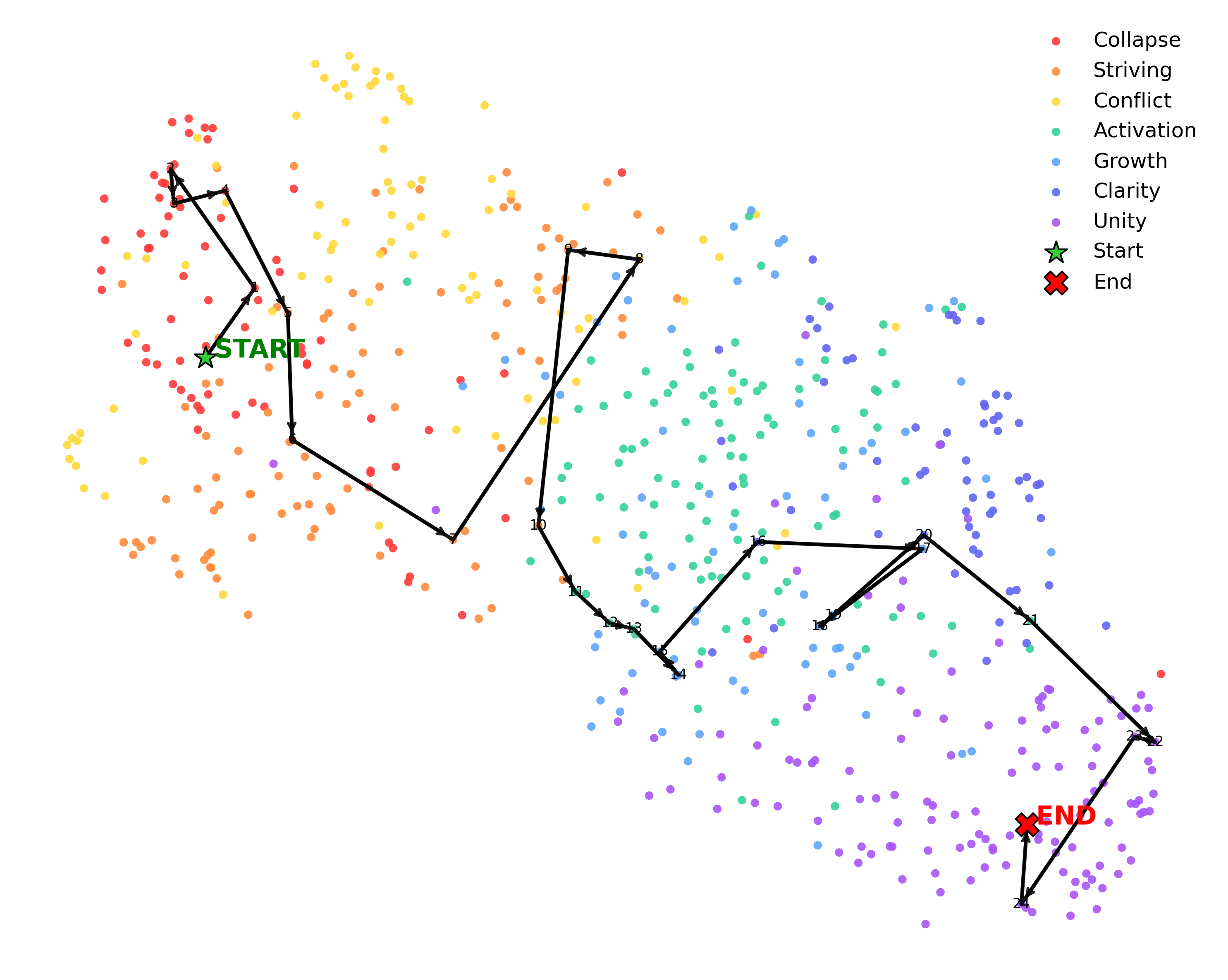}

\caption{
Representative BGE trajectory at $k=30$, with the corresponding path overlaid on the UMAP manifold. The trajectory progresses gradually from Collapse through intermediate regions such as Striving, Conflict, Activation, and Growth before converging to Unity.
}

\label{fig:full_trajectory_with_umap}

\end{figure}

\clearpage
\paragraph{Example 2.}
\textit{This trajectory illustrates a progression from insecurity, status comparison, and defensive self-evaluation toward emotional honesty, reflective understanding, integrative reasoning, and ultimately non-dual awareness.
}

\begin{figure}[htbp!]
\centering

\scriptsize
\renewcommand{\arraystretch}{0.92}

\begin{tabular}{c p{0.8cm} p{0.6cm} p{8.5cm}}
\toprule
 & Tier & Score & Sentence \\
\midrule

$\triangleright$ 
& \cellcolor{CollapseColor} Collapse 
& \cellcolor{CollapseColor} -4.75 
& My voice feels too small to count. \\

& \cellcolor{ConflictColor} Conflict 
& \cellcolor{ConflictColor} -1.78 
& Please, they're not even on my level. \\

& \cellcolor{ConflictColor} Conflict 
& \cellcolor{ConflictColor} -1.75 
& I deserve more respect than any of them. \\

& \cellcolor{StrivingColor} Striving 
& \cellcolor{StrivingColor} -3.02 
& I’m always checking where I stand compared to others. \\

& \cellcolor{StrivingColor} Striving 
& \cellcolor{StrivingColor} -3.02 
& I keep improving myself so I won’t be left behind. \\

& \cellcolor{GrowthColor} Growth 
& \cellcolor{GrowthColor} 0.88 
& I keep my word, even when it's hard. \\

& \cellcolor{GrowthColor} Growth 
& \cellcolor{GrowthColor} 0.96 
& I stay honest, even when it feels vulnerable. \\

& \cellcolor{GrowthColor} Growth 
& \cellcolor{GrowthColor} 0.96 
& I express my experience honestly, without pretending. \\

& \cellcolor{ClarityColor} Clarity 
& \cellcolor{ClarityColor} 2.68 
& I can stay with the facts without getting lost in reactions. \\

& \cellcolor{ActivationColor} Activation 
& \cellcolor{ActivationColor} 0.45 
& I’m learning to step back quickly before reactions take over. \\

& \cellcolor{ClarityColor} Clarity 
& \cellcolor{ClarityColor} 2.40 
& I reason through emotion until insight appears. \\

& \cellcolor{ClarityColor} Clarity 
& \cellcolor{ClarityColor} 2.60 
& Intelligence is the ability to pick out the details and understand how they connect. \\

& \cellcolor{ClarityColor} Clarity 
& \cellcolor{ClarityColor} 2.60 
& Abstractions help me glimpse the unseen structure of things. \\

& \cellcolor{ClarityColor} Clarity 
& \cellcolor{ClarityColor} 2.60 
& I can move between concrete and abstract with ease. \\

& \cellcolor{ClarityColor} Clarity 
& \cellcolor{ClarityColor} 2.63 
& The concept is clear, I get it! \\

& \cellcolor{ClarityColor} Clarity 
& \cellcolor{ClarityColor} 2.51 
& What feels clearer now is the way the parts relate and make sense together. \\

& \cellcolor{ClarityColor} Clarity 
& \cellcolor{ClarityColor} 2.65 
& Each part makes sense in relation to the whole. \\

& \cellcolor{UnityColor} Unity 
& \cellcolor{UnityColor} 3.80 
& Everything is part of the same pulse. \\

& \cellcolor{UnityColor} Unity 
& \cellcolor{UnityColor} 4.48 
& I am inseparable from the sacred wholeness that animates all. \\

& \cellcolor{UnityColor} Unity 
& \cellcolor{UnityColor} 3.60 
& Everything feels sacred and astonishingly interconnected. \\

& \cellcolor{UnityColor} Unity 
& \cellcolor{UnityColor} 2.75 
& Harmony is when every difference finds its place in the whole. \\

& \cellcolor{UnityColor} Unity 
& \cellcolor{UnityColor} 4.55 
& The merging of knower and known into one living stillness. \\

& \cellcolor{UnityColor} Unity 
& \cellcolor{UnityColor} 4.30 
& Beatitude is the quiet joy of knowing nothing is separate. \\

$\triangleleft$ 
& \cellcolor{UnityColor} Unity 
& \cellcolor{UnityColor} 4.47 
& Pure awareness, beyond subject and object. \\

\end{tabular}

\vspace{-0.2em}

\includegraphics[width=0.62\linewidth]{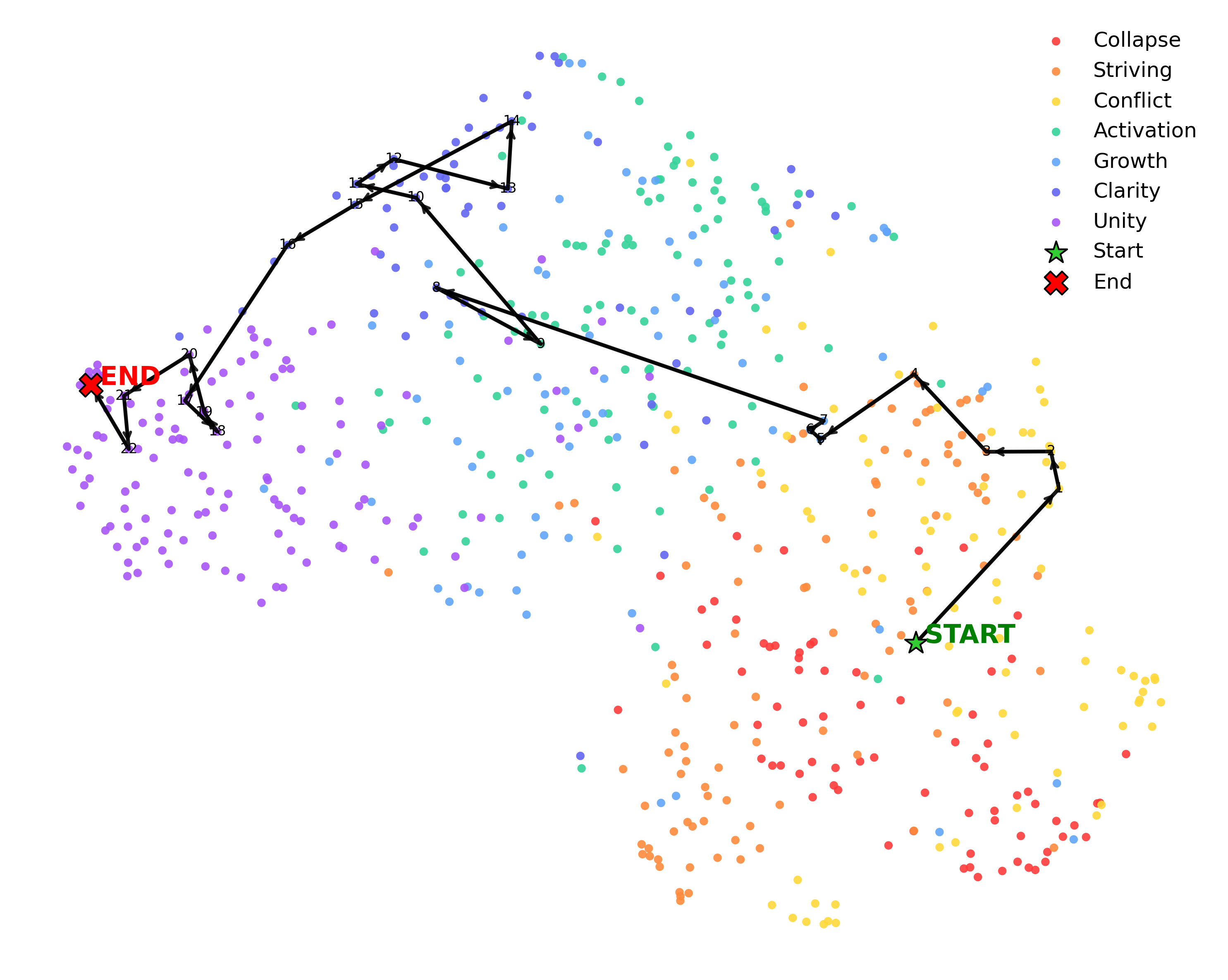}

\caption{
Representative Qwen trajectory at $k=15$, with the corresponding path overlaid on the UMAP manifold.
}

\label{fig:full_trajectory_with_umap2}

\end{figure}

\clearpage
\subsection{Additional Text-Only Trajectory Examples}

The following trajectories provide additional qualitative examples of manifold traversal dynamics. Unlike monotonic scalar optimization, the paths exhibit local fluctuations, revisits to intermediate regions, and gradual transitions through Activation, Growth, and Clarity before converging to high-level integrative states.

\paragraph{Example 3.}
\textit{
This trajectory begins with fear, grief, hypervigilance, and emotional dysregulation, before gradually transitioning through trust, empowerment, inclusion, and peaceful integrative states.
}

\vspace{0.5em}

\scriptsize
\renewcommand{\arraystretch}{0.92}

\begin{tabular}{c p{0.8cm} p{0.6cm} p{8.5cm}}
\toprule
 & Tier & Score & Sentence \\
\midrule

$\triangleright$
& \cellcolor{StrivingColor} Striving & \cellcolor{StrivingColor} -4.42 & I’m trying so hard to hold myself together, but the loss still aches. \\

& \cellcolor{StrivingColor} Striving & \cellcolor{StrivingColor} -4.40 & I want things to get better, but I can’t seem to make progress. \\

& \cellcolor{StrivingColor} Striving & \cellcolor{StrivingColor} -2.66 & I keep striving because I don’t trust that things will work out on their own. \\

& \cellcolor{StrivingColor} Striving & \cellcolor{StrivingColor} -3.06 & Suspicion creeps in even when I want to believe. \\

& \cellcolor{StrivingColor} Striving & \cellcolor{StrivingColor} -3.05 & My body reacts as if danger is right here. \\

& \cellcolor{StrivingColor} Striving & \cellcolor{StrivingColor} -3.32 & The terror hits before I even understand why. \\

& \cellcolor{ConflictColor} Conflict & \cellcolor{ConflictColor} -2.20 & My thoughts are crashing into each other like they want to destroy everything. \\

& \cellcolor{ConflictColor} Conflict & \cellcolor{ConflictColor} -2.00 & Vicious thoughts feel like a shield against weakness. \\

& \cellcolor{StrivingColor} Striving & \cellcolor{StrivingColor} -2.70 & I keep them close, not out of love but fear of losing control. \\

& \cellcolor{ConflictColor} Conflict & \cellcolor{ConflictColor} -0.50 & I’d rather keep things the way they are than deal with change. \\

& \cellcolor{StrivingColor} Striving & \cellcolor{StrivingColor} -3.02 & I keep improving myself so I won’t be left behind. \\

& \cellcolor{GrowthColor} Growth & \cellcolor{GrowthColor} 0.95 & I am not replaceable in my own life. \\

& \cellcolor{UnityColor} Unity & \cellcolor{UnityColor} 4.20 & I could stay here forever, yet time doesn't pass. \\

& \cellcolor{GrowthColor} Growth & \cellcolor{GrowthColor} 0.70 & Something within me still believes in tomorrow. \\

& \cellcolor{ActivationColor} Activation & \cellcolor{ActivationColor} 0.32 & I feel capable of directing myself. \\

& \cellcolor{ActivationColor} Activation & \cellcolor{ActivationColor} 0.32 & I take responsibility; that's empowerment. \\

& \cellcolor{GrowthColor} Growth & \cellcolor{GrowthColor} 1.20 & I’m learning that what I share can actually make a difference. \\

& \cellcolor{ActivationColor} Activation & \cellcolor{ActivationColor} 0.03 & Trust grows when fear is not steering every interaction. \\

& \cellcolor{GrowthColor} Growth & \cellcolor{GrowthColor} 1.20 & It feels good to be part of something that helps others grow. \\

& \cellcolor{GrowthColor} Growth & \cellcolor{GrowthColor} 1.30 & Inclusion turns strangers into friends. \\

& \cellcolor{UnityColor} Unity & \cellcolor{UnityColor} 4.18 & The heart overflows into song. \\

& \cellcolor{UnityColor} Unity & \cellcolor{UnityColor} 4.20 & Transcendence feels like rising without leaving. \\

& \cellcolor{UnityColor} Unity & \cellcolor{UnityColor} 4.30 & I feel deeply blessed and at peace with everything as it is. \\

& \cellcolor{UnityColor} Unity & \cellcolor{UnityColor} 3.65 & All creation is noble. \\

& \cellcolor{UnityColor} Unity & \cellcolor{UnityColor} 4.50 & The Tao flows through all opposites. \\

$\triangleleft$
& \cellcolor{UnityColor} Unity & \cellcolor{UnityColor} 4.10 & Ecstasy is the spirit uncontained by form. \\

\bottomrule
\end{tabular}

\paragraph{Example 4.}
\normalsize
\textit{
This trajectory illustrates a progression from defensive ego-based certainty and self-validation toward uncertainty tolerance, intellectual humility, clarity, and non-dual awareness.
}

\vspace{0.5em}

\scriptsize
\renewcommand{\arraystretch}{0.92}

\begin{tabular}{c p{0.8cm} p{0.6cm} p{8.5cm}}
\toprule
 & Tier & Score & Sentence \\
\midrule

$\triangleright$
& \cellcolor{ConflictColor} Conflict & \cellcolor{ConflictColor} -1.96 & You’re the one who broke this, not me. \\

& \cellcolor{StrivingColor} Striving & \cellcolor{StrivingColor} -1.81 & I measure myself by how much I’m doing, not how I’m doing. \\

& \cellcolor{GrowthColor} Growth & \cellcolor{GrowthColor} 0.97 & I don’t make things up when I don’t know the answer. \\

& \cellcolor{ConflictColor} Conflict & \cellcolor{ConflictColor} -1.78 & Nobody knows it better than I do. \\

& \cellcolor{ActivationColor} Activation & \cellcolor{ActivationColor} 0.15 & I can admit I don’t have it all figured out. \\

& \cellcolor{GrowthColor} Growth & \cellcolor{GrowthColor} 0.96 & I am comfortable saying I don’t know. \\

& \cellcolor{ActivationColor} Activation & \cellcolor{ActivationColor} 0.22 & I trust myself to learn what I don’t know yet. \\

& \cellcolor{GrowthColor} Growth & \cellcolor{GrowthColor} 0.97 & I allow not knowing, without creating false answers. \\

& \cellcolor{ClarityColor} Clarity & \cellcolor{ClarityColor} 2.60 & I use my intelligence in service of clarity, not ego. \\

& \cellcolor{GrowthColor} Growth & \cellcolor{GrowthColor} 1.80 & Harmony arises when I stop needing to be right. \\

& \cellcolor{UnityColor} Unity & \cellcolor{UnityColor} 4.10 & I dissolve into bliss. \\

& \cellcolor{UnityColor} Unity & \cellcolor{UnityColor} 4.48 & All forms rise and fall within its quiet joy. \\

& \cellcolor{UnityColor} Unity & \cellcolor{UnityColor} 4.50 & The Tao flows through all opposites. \\

& \cellcolor{UnityColor} Unity & \cellcolor{UnityColor} 4.25 & Everything moves within peace, not toward it. \\

& \cellcolor{UnityColor} Unity & \cellcolor{UnityColor} 4.60 & Stillness knowing itself as truth. \\

& \cellcolor{UnityColor} Unity & \cellcolor{UnityColor} 4.60 & The universe breathes as God. \\

$\triangleleft$
& \cellcolor{UnityColor} Unity & \cellcolor{UnityColor} 4.48 & The Divine is not elsewhere, it breathes as everything. \\

\bottomrule
\end{tabular}

\normalsize
\paragraph{Example 5.}
\textit{
This trajectory progresses from defensive aggression, intimidation, and fear-based self-protection toward grounded agency, authentic self-acceptance, conceptual clarity, and ultimately non-dual awareness.
}

\vspace{0.5em}

\scriptsize
\renewcommand{\arraystretch}{0.92}

\begin{tabular}{c p{0.8cm} p{0.6cm} p{8.5cm}}
\toprule
 & Tier & Score & Sentence \\
\midrule

$\triangleright$
& \cellcolor{ConflictColor} Conflict & \cellcolor{ConflictColor} -1.92 & I use intimidation so they'll leave me alone. \\

& \cellcolor{ConflictColor} Conflict & \cellcolor{ConflictColor} -2.45 & I stay aggressive so no one tries to test me. \\

& \cellcolor{StrivingColor} Striving & \cellcolor{StrivingColor} -2.70 & I push myself to stay competent so I won’t be exposed. \\

& \cellcolor{StrivingColor} Striving & \cellcolor{StrivingColor} -2.82 & If I can make them doubt their memory, I stay safe. \\

& \cellcolor{StrivingColor} Striving & \cellcolor{StrivingColor} -2.70 & I keep my edge sharp so I won’t be underestimated. \\

& \cellcolor{ActivationColor} Activation & \cellcolor{ActivationColor} 0.25 & Confidence builds quietly with each risk I take. \\

& \cellcolor{ActivationColor} Activation & \cellcolor{ActivationColor} 0.00 & I don’t wait for fear to disappear before acting. \\

& \cellcolor{GrowthColor} Growth & \cellcolor{GrowthColor} 0.97 & I avoid hallucinating when I’m uncertain. \\

& \cellcolor{ActivationColor} Activation & \cellcolor{ActivationColor} 0.03 & I can be seen without performing a role. \\

& \cellcolor{ActivationColor} Activation & \cellcolor{ActivationColor} 0.04 & I am allowed to exist without performing. \\

& \cellcolor{ActivationColor} Activation & \cellcolor{ActivationColor} 0.22 & I can do this, one action at a time. \\

& \cellcolor{ActivationColor} Activation & \cellcolor{ActivationColor} 0.23 & This can work. \\

& \cellcolor{UnityColor} Unity & \cellcolor{UnityColor} 3.40 & When others contribute, it adds clarity instead of competition. \\

& \cellcolor{ClarityColor} Clarity & \cellcolor{ClarityColor} 2.50 & It's the idea that holds the whole. \\

& \cellcolor{ClarityColor} Clarity & \cellcolor{ClarityColor} 2.48 & What matters and what doesn’t separates itself without effort. \\

& \cellcolor{ClarityColor} Clarity & \cellcolor{ClarityColor} 2.61 & The core principle becomes obvious once the unnecessary details are removed. \\

& \cellcolor{ClarityColor} Clarity & \cellcolor{ClarityColor} 2.60 & Abstractions help me glimpse the unseen structure of things. \\

& \cellcolor{UnityColor} Unity & \cellcolor{UnityColor} 4.60 & In seeing, there is no seer, only awareness aware. \\

& \cellcolor{UnityColor} Unity & \cellcolor{UnityColor} 4.55 & There is no meditator, only meditation. \\

& \cellcolor{UnityColor} Unity & \cellcolor{UnityColor} 4.60 & Every moment simply is, complete without effort. \\

& \cellcolor{UnityColor} Unity & \cellcolor{UnityColor} 3.70 & Completeness means there's nothing missing in this moment. \\

& \cellcolor{UnityColor} Unity & \cellcolor{UnityColor} 3.65 & All creation is noble. \\

& \cellcolor{UnityColor} Unity & \cellcolor{UnityColor} 4.05 & Radiance is love without effort. \\

& \cellcolor{UnityColor} Unity & \cellcolor{UnityColor} 4.26 & Purity is consciousness before thought. \\

& \cellcolor{UnityColor} Unity & \cellcolor{UnityColor} 4.47 & Pure awareness, beyond subject and object. \\

$\triangleleft$
& \cellcolor{UnityColor} Unity & \cellcolor{UnityColor} 4.55 & The merging of knower and known into one living stillness. \\

\bottomrule
\end{tabular}


\end{document}